%% file: 0-0KDD-authordraft.tex
\documentclass[sigconf]{acmart}

\AtBeginDocument{%
  \providecommand\BibTeX{{%
    \normalfont B\kern-0.5em{\scshape i\kern-0.25em b}\kern-0.8em\TeX}}}

\copyrightyear{2023}
\acmYear{2023}
\setcopyright{rightsretained}
\acmConference[KDD '23]{Proceedings of the 29th ACM SIGKDD Conference on Knowledge Discovery and Data Mining}{August 6--10, 2023}{Long Beach, CA, USA}
\acmBooktitle{Proceedings of the 29th ACM SIGKDD Conference on Knowledge Discovery and Data Mining (KDD '23), August 6--10, 2023, Long Beach, CA, USA}
\acmDOI{10.1145/3580305.3599485}
\acmISBN{979-8-4007-0103-0/23/08}

\acmSubmissionID{rtfp1392}

\usepackage{soul}
\usepackage{graphicx}
\usepackage{amsmath}
\usepackage{amsthm}
\usepackage{booktabs}
\usepackage{algorithm}
\usepackage{algorithmic}
\usepackage[algo2e]{algorithm2e}
\usepackage{mathtools}     

\usepackage{subcaption}     

\usepackage{subfiles}     
\newcommand{\tool}{\sc Rapid}

\makeatletter
\gdef\@copyrightpermission{
  \begin{minipage}{0.3\columnwidth}
   \href{https://nam04.safelinks.protection.outlook.com/?url=https\%3A\%2F\%2Fcreativecommons.org\%2Flicenses\%2Fby\%2F4.0\%2F&data=05\%7C01\%7Ctu85\%40purdue.edu\%7C4deb4948d7d741001fdd08db6e68dd8e\%7C4130bd397c53419cb1e58758d6d63f21\%7C0\%7C0\%7C638225169250264567\%7CUnknown\%7CTWFpbGZsb3d8eyJWIjoiMC4wLjAwMDAiLCJQIjoiV2luMzIiLCJBTiI6Ik1haWwiLCJXVCI6Mn0\%3D\%7C3000\%7C\%7C\%7C&sdata=EbhMBw71WMKww7CtI\%2F6sjazkTZQUyTWaV8DdQruIC\%2Bc\%3D&reserved=0}{\includegraphics[width=0.90\textwidth]{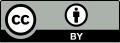}}
  \end{minipage}\hfill
  \begin{minipage}{0.7\columnwidth}
   \href{https://nam04.safelinks.protection.outlook.com/?url=https\%3A\%2F\%2Fcreativecommons.org\%2Flicenses\%2Fby\%2F4.0\%2F&data=05\%7C01\%7Ctu85\%40purdue.edu\%7C4deb4948d7d741001fdd08db6e68dd8e\%7C4130bd397c53419cb1e58758d6d63f21\%7C0\%7C0\%7C638225169250264567\%7CUnknown\%7CTWFpbGZsb3d8eyJWIjoiMC4wLjAwMDAiLCJQIjoiV2luMzIiLCJBTiI6Ik1haWwiLCJXVCI6Mn0\%3D\%7C3000\%7C\%7C\%7C&sdata=EbhMBw71WMKww7CtI\%2F6sjazkTZQUyTWaV8DdQruIC\%2Bc\%3D&reserved=0}{This work is licensed under a Creative Commons Attribution International 4.0 License.}
  \end{minipage}
  \vspace{5pt}
}
\makeatother

\begin{document}

\title{Rapid Image Labeling via Neuro-Symbolic Learning}

\author{Yifeng Wang}
\email{yifeng.wang@connect.polyu.hk}
\affiliation{%
  \institution{The Hong Kong Polytechnic University}
  \city{Hong Kong}
  \country{China}
}

\authornote{Both authors contributed equally to this research.}
\authornote{This work was done when these students were research interns at Purdue University.}

\author{Zhi Tu}
\authornotemark[1]
\email{tu85@purdue.edu}
\affiliation{%
  \institution{Purdue University}
  \city{West Lafayette}
  \country{USA}
}

\author{Yiwen Xiang}
\authornotemark[2]
\email{20183749@cqu.edu.cn}
\affiliation{%
  \institution{Chongqing University}
  \city{Chongqing}
  \country{China}
}

\author{Shiyuan Zhou}
\authornotemark[2]
\email{shiyuan.zhou@mail.utoronto.ca}
\affiliation{%
  \institution{University of Toronto}
  \city{Toronto}
  \country{Canada}
}

\author{Xiyuan Chen}
\authornotemark[2]
\email{garethcxy.chen@mail.utoronto.ca}
\affiliation{%
 \institution{University of Toronto}
 \city{Toronta}
 \country{Canada}
}

\author{Bingxuan Li}
\email{li3393@purdue.edu}
\affiliation{%
  \institution{Purdue University}
  \city{West Lafayette}
  \country{USA}
}

\author{Tianyi Zhang}
\email{tianyi@purdue.edu}
\affiliation{%
  \institution{Purdue University}
  \city{West Lafayette}
  \country{USA}
}

\renewcommand{\shortauthors}{Yifeng Wang et al.}

\subfile{0-Abstract.tex}



\begin{teaserfigure}
\centering
\includegraphics[width=0.83\textwidth]{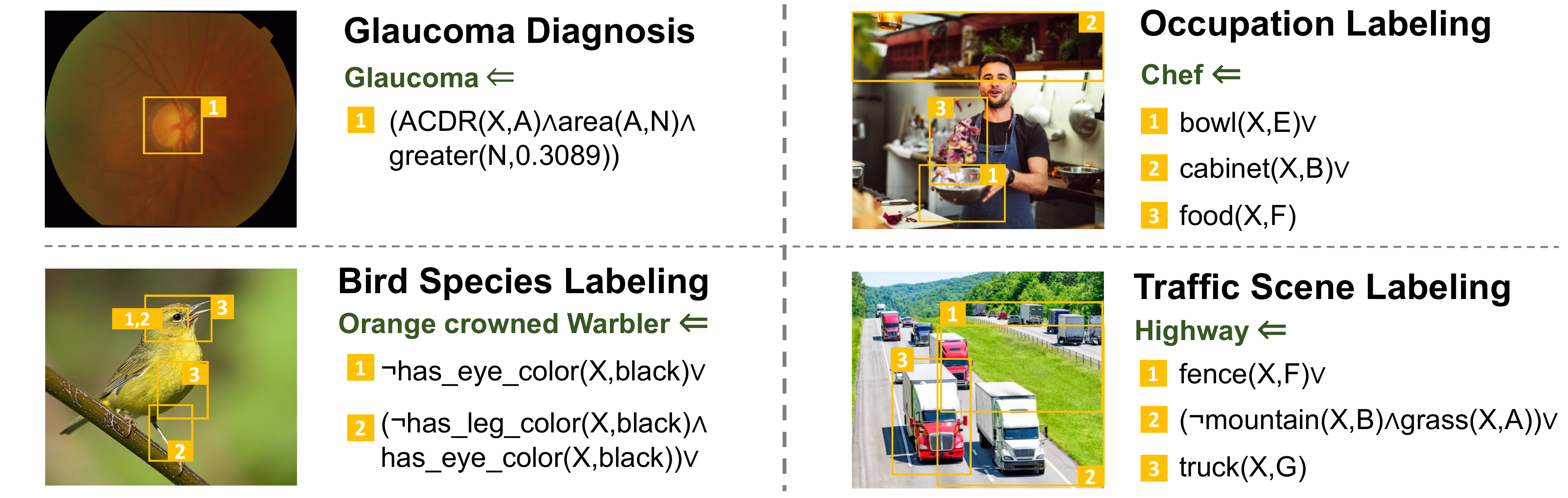}
  \caption{Two image labeling tasks from highly specialized domains and two from common domains with example labeling rules.}
  \label{fig:teaser}
\end{teaserfigure}

\received{20 February 2007}
\received[revised]{12 March 2009}
\received[accepted]{5 June 2009}

\maketitle

\subfile{1-Introduction.tex}

\subfile{2-RelatedWork.tex}

\subfile{3-Method.tex}

\subfile{4-Experiments.tex}

\subfile{5-Discussion.tex}

\subfile{6-Conclusion.tex}

\section*{Acknowledgement}
We would like to thank the anonymous reviewers for their valuable feedback. This work was in part supported by an Amazon Research Award.




\bibliographystyle{ACM-Reference-Format}
\bibliography{mybibliography}

\appendix

\subfile{7-Appendix.tex}

\end{document}


\title{Appendix}

\author{Ben Trovato}
\authornote{Both authors contributed equally to this research.}
\email{trovato@corporation.com}
\orcid{1234-5678-9012}
\author{G.K.M. Tobin}
\authornotemark[1]
\email{webmaster@marysville-ohio.com}
\affiliation{%
  \institution{Institute for Clarity in Documentation}
  \streetaddress{P.O. Box 1212}
  \city{Dublin}
  \state{Ohio}
  \country{USA}
  \postcode{43017-6221}
}

\author{Lars Th{\o}rv{\"a}ld}
\affiliation{%
  \institution{The Th{\o}rv{\"a}ld Group}
  \streetaddress{1 Th{\o}rv{\"a}ld Circle}
  \city{Hekla}
  \country{Iceland}}
\email{larst@affiliation.org}

\author{Valerie B\'eranger}
\affiliation{%
  \institution{Inria Paris-Rocquencourt}
  \city{Rocquencourt}
  \country{France}
}

\author{Aparna Patel}
\affiliation{%
 \institution{Rajiv Gandhi University}
 \streetaddress{Rono-Hills}
 \city{Doimukh}
 \state{Arunachal Pradesh}
 \country{India}}

\author{Huifen Chan}
\affiliation{%
  \institution{Tsinghua University}
  \streetaddress{30 Shuangqing Rd}
  \city{Haidian Qu}
  \state{Beijing Shi}
  \country{China}}

\author{Charles Palmer}
\affiliation{%
  \institution{Palmer Research Laboratories}
  \streetaddress{8600 Datapoint Drive}
  \city{San Antonio}
  \state{Texas}
  \country{USA}
  \postcode{78229}}
\email{cpalmer@prl.com}

\author{John Smith}
\affiliation{%
  \institution{The Th{\o}rv{\"a}ld Group}
  \streetaddress{1 Th{\o}rv{\"a}ld Circle}
  \city{Hekla}
  \country{Iceland}}
\email{jsmith@affiliation.org}

\author{Julius P. Kumquat}
\affiliation{%
  \institution{The Kumquat Consortium}
  \city{New York}
  \country{USA}}
\email{jpkumquat@consortium.net}

\renewcommand{\shortauthors}{Trovato and Tobin, et al.}

\subfile{0-Abstract.tex}

\ccsdesc[500]{Computer systems organization~Embedded systems}
\ccsdesc[300]{Computer systems organization~Redundancy}
\ccsdesc{Computer systems organization~Robotics}
\ccsdesc[100]{Networks~Network reliability}

\keywords{datasets, neural networks, gaze detection, text tagging}


\received{20 February 2007}
\received[revised]{12 March 2009}
\received[accepted]{5 June 2009}

\maketitle

\bibliographystyle{ACM-Reference-Format}
\bibliography{mybibliography}

\appendix

%% file: 0-Abstract.tex
\begin{abstract}
The success of Computer Vision (CV) relies heavily on manually annotated data. 
However, it is prohibitively expensive to annotate images in key domains such as healthcare, where data labeling requires significant domain expertise and cannot be easily delegated to crowd workers. 
To address this challenge, we propose a neuro-symbolic approach called {\tool}, which infers image labeling rules from a small amount of labeled data provided by domain experts 
and automatically labels unannotated data using the rules.
Specifically, {\tool} combines pre-trained CV models and inductive logic learning to infer the logic-based labeling rules.
{\tool} achieves a labeling accuracy of $83.33\%$ to $88.33\%$ on four image labeling tasks with only 12 to 39 labeled samples. In particular, {\tool} significantly outperforms finetuned CV models  in two highly specialized tasks. These results demonstrate the effectiveness of {\tool} in learning from small data and its capability to generalize among different tasks.
Code and our dataset are publicly available at \href{https://github.com/Neural-Symbolic-Image-Labeling/Rapid/}{https://github.com/Neural-Symbolic-Image-Labeling/Rapid/}
\end{abstract}

%% file: 1-Introduction.tex
\section{Introduction}
\label{sec:intro}
Deep learning methods have shown great power in challenging computer vision tasks, such as traffic scene detection~\cite{01,02} and disease diagnosis~\cite{03,04}. 
These methods often require a vast amount of labeled image data to achieve good performance. Labeling these data is laborious and expensive. This challenge is exacerbated in highly specialized domains such as healthcare, where  data labeling requires significant domain expertise.

Many data labeling methods have been proposed to address this challenge with a small amount of labeled data (e.g., less than 100 labeled samples).
One mainstream line of research is to develop models for automated data labeling \cite{16,18,14,titano2018automated}.
Existing approaches in this category often learn class prototypes from the training data samples and infer the class of unlabeled data by assigning the class of its nearest class prototype. 
To adopt these approaches in a low resource setting, the distance between data samples is often designed to depend on task-specific information such as meta-data or other task-specific insights.
However, the task-specific nature of these approaches restricts their generalizability to other tasks.
Besides, the need for designing specific models for a specific task requires extensive human efforts, which is against the motivation of saving human efforts for data labeling in the first place.

To address the aforementioned limitations, a new data labeling paradigm called data programming~\cite{DBLP:conf/nips/RatnerSWSR16} has been proposed for rapid data labeling. For example, Snorkel~\cite{ratner2017snorkel} asks domain experts to create labeling functions and uses a generative model to combine those labeling functions to provide probabilistic labels.
However, those labeling functions must be written in a programming language such as Python.
This requirement not only incurs an overhead of manually composing labeling functions but also incurs a steep learning curve for domain experts and end users who typically do not have any programming experience. 

In this paper, we propose a new neuro-symbolic approach called {\tool} for image labeling in low-resource settings (e.g., less than 100 labeled images). 
The novelty of this framework lies in synergizing the strength of neural models (i.e., handling rich, complex image data) and the strength of inductive logic learning (i.e., learning from small datasets) to handle the image labeling challenge. 
Specifically, {\tool} automatically infers logic rules from a small amount of labeled data, applies the inferred rules to label the unlabeled data, and solicits user feedback to refine the rules iteratively.

Unlike Snorkel, {\tool} infers labeling rules automatically rather than requiring users to manually construct these rules. {\tool} leverages the First-Order Inductive Learner  (FOIL) algorithm to infer logic rules based on the low-level visual attributes extracted by pre-trained models. This way, our approach disentangles the perception and the learning process, making it more transparent and explainable to human labelers. Furthermore, to maximize the efficiency of data usage, we develop a multi-criteria active learning method to iteratively elicit human feedback to refine the labeling rules.

We conduct extensive experiments on datasets from two highly specialized domains and two common domains. 
Our method achieves significantly higher labeling accuracy on the two highly specialized domains ($85.52\%$ on disease diagnosis and $86.11\%$ on bird species labeling, respectively) compared to the baseline models.
We demonstrate that by actively refining labeling rules with rapid, incremental human feedback, {\tool} can effectively embed expert knowledge and achieve high image labeling accuracy with a limited amount of training data. 


Overall, this work makes the following contributions: 
\begin{itemize}
    \item We proposed a new neuro-symbolic learning framework that synergizes pre-trained computer vision models with inductive logic learning for rapid image labeling with a limited amount of training data;
    \item We designed a new conflict-based informativeness metric for data selection in active learning;
    \item We conducted comprehensive evaluations on four labeling tasks from different domains with user simulation and multiple baselines.
\end{itemize}

%% file: 2-RelatedWork.tex
\section{Related Work}
\label{sec:related}
\subsection{Image Labeling}
Reducing the human effort in image labeling has become increasingly important in recent years with the advent of deep learning,  which requires a massive amount of labeled data to train a model. There has been a continuous effort to develop automated solutions for image labeling.
The basic idea is to train a model with some labeled data and automatically assign labels to new data samples without further human involvement.
Among these existing methods, fully-automated methods have attracted significant attention and achieved promising performance.
Some methods exploit the similarity between unlabeled and label images~\cite{16}.
Some methods resolve the problem of data scarcity by creating representations of images using auxiliary information such as corresponding captions~\cite{18}, meta-data~\cite{14}, or pseudo-labels generated by other models~\cite{titano2018automated}.
Despite the promising performance of these methods, they often lack generalizability to domains where data acquisition is challenging, as the models usually require a substantial amount of data to learn the knowledge.
Thus, some semi-automated labeling approaches use human efforts in the training or inference process to provide the information needed for training~\cite{suchi2019easylabel} or a coarse initial label~\cite{grady2004multi}.
Compared with existing work, our approach uses inductive logic learning to infer logic rules from a small amount of human-annotated data to label images.

\subsection{Active Learning}
Active learning is widely adopted to get humans involved in the labeling process iteratively while minimizing the amount of data to be labeled by humans.
Active learning methods select the data samples that can benefit the model most in each iteration and request humans to label them to push the usage of human efforts to the minimum. 
To determine which data sample to be labeled by humans, some approaches use probability models
and prioritize the data samples with high prediction inconsistency~\cite{zhang2008active,gao2020consistency,guo2021semi,9880354}, 
some depend on the vectorized representation from deep learning models~\cite{ho2020deep,9009538,zhang2022boostmis},
and some calculate the low-rank matrix representation for both labeled and unlabeled data to calculate the informativeness~\cite{wu2017adaptive}.

Though the active learning strategy can optimize data selection to reduce human effort, it often requires many iterations to achieve a reasonable labeling accuracy. Thus, it remains too expensive for labeling tasks where time is precious for domain experts such as clinicians. 
Snorkel~\cite{ratner2017snorkel} enables users to explicitly embed their domain knowledge by creating labeling functions.
However, the labeling functions are either written in programming languages or special declarative functions defined by the author of Snorkel, causing huge overhead effort in learning the labeling function grammar.
Our work combines interactive learning with an inductive logic learner, generating logic rules to classify images. 
The generated rules are represented with simple logic and descriptive predicates, which are easy for users to read and edit.

\subsection{Neuro-Symbolic Learning}
There has been a growing interest in combining neural networks with symbolic methods~\cite{DBLP:conf/iclr/MaoGKTW19,yi2018neural,
hudson2019learning,DBLP:conf/icml/AmizadehPPHK20,DBLP:conf/ijcai/MuraliSKM22}.
Here, the term {\em neuro} refers to artificial neural networks or connectionist systems, while the term {\em symbolic} refers to AI approaches that perform explicit symbol manipulation, such as term rewriting, graph algorithms, and formal logic. There are different ways of combining neural network modules with symbolic learning. Following the categorization in a recent survey~\cite{DBLP:journals/corr/abs-2105-05330}, 
our method belongs to a cascading neuro-symbolic paradigm that extracts latent patterns from input data using a neural system and then feeds them into a symbolic reasoner for final prediction. 

Existing approaches in this category include NS-VQA~\cite{NEURIPS2018_5e388103}, NS-CL~\cite{DBLP:conf/iclr/MaoGKTW19}, and FO-SL~\cite{DBLP:conf/ijcai/MuraliSKM22}.
NS-VQA~\cite{NEURIPS2018_5e388103} firstly parses an image to a structural scene representation with Mask R-CNN and ResNet-34 and converts a natural language question into a query program with an LSTM model. Then it uses a symbolic executor to run the program on the scene representation to obtain the answer to the given question.
NS-CL~\cite{DBLP:conf/iclr/MaoGKTW19} adopts a similar approach as NL-VQA but learns the feature vector representation of an object from question-answer pairs, instead of extracting them directly with pre-trained models.
FO-SL~\cite{DBLP:conf/ijcai/MuraliSKM22} represents images in first-order logic and uses an SAT solver to solve visual discrimination puzzles.



Unlike existing neuro-symbolic learning approaches in this category, we are the first to use inductive logic learning as the symbolic method for rule inference. In this way, we can explicitly model the logic of labeling rules. Furthermore, our approach is also the first to apply neuro-symbolic learning to image labeling.

\subsection{Human Feedback in Inductive Logic Learning}
To the best of our knowledge, our work is the first that integrates active learning with Inductive Logic Learning (ILL). Existing research in ILL has focused on improving the learning algorithm for better efficiency and scalability. There are only a few that investigates the interactivity of ILL in the 1990s \cite{de1992overview, de1992interactive, bergadano1993interactive}. Specifically, De Raedt et al.~\cite{de1992interactive, de1992overview} propose an interactive paradigm in which the inductive learner asks a yes/no question about the correctness of a learned rule. If a user answers no, the learner will backtrack and learn a new rule. Bergadano et al.~\cite{bergadano1993interactive} propose to prompt users for new counter-examples to refute an incorrect logic program, but users need to manually design counter-examples from scratch. More recently, Sivaraman et al.~\cite{sivaraman2019active} present an interactive inductive logic programming approach to infer rule-based code patterns for code search. However, this approach only allows users to mark some search results as correct or incorrect for pattern refinement. None of these four approaches have an active learning component (i.e., a data selection algorithm) to carefully rank and select data samples for user inspection and correction.

%% file: 3-Method.tex
\begin{figure}[ht]
\begin{center}
 \includegraphics[width=\linewidth]{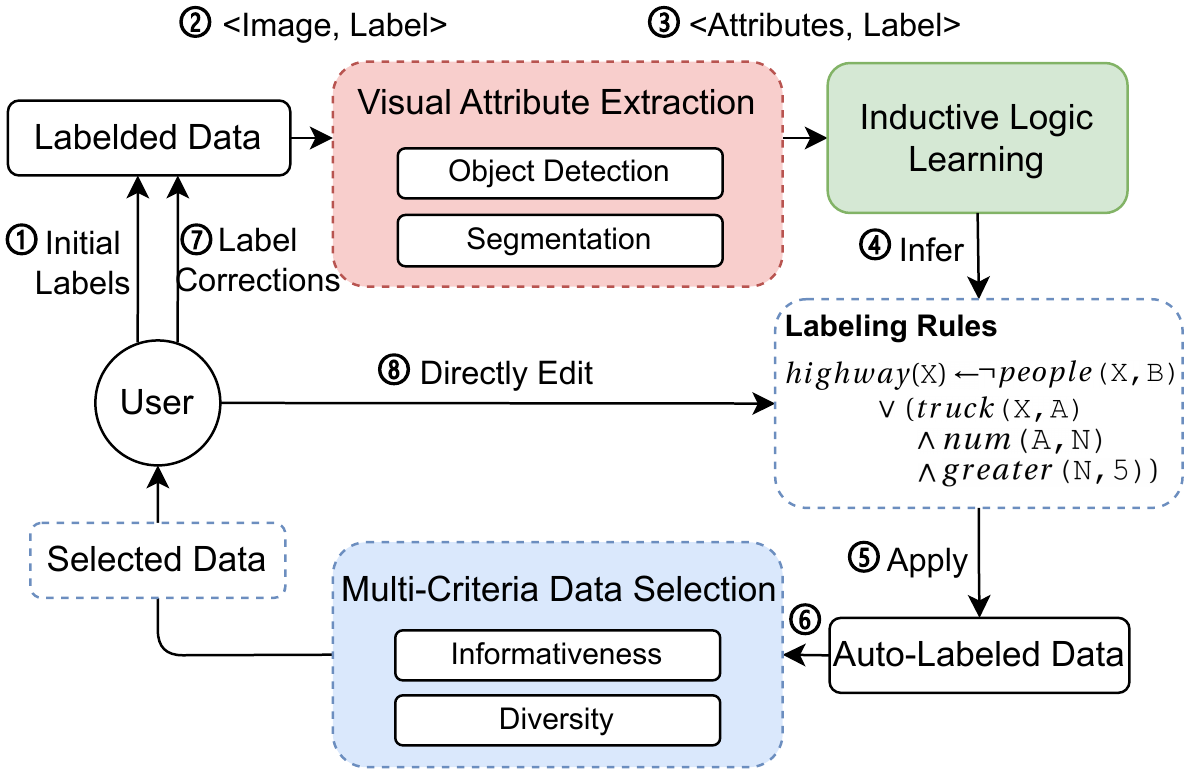}
 \caption{The overview of our image labeling approach.}
 \label{fig:pipeline}
\end{center}
\end{figure}

\section{Method}
Figure~\ref{fig:pipeline} gives an overview of our approach called {\tool}. {\tool} consists of three parts---(1) a pre-trained visual attribute extractor to extract basic, low-level visual attributes from images, (2) an inductive learner to infer logic labeling rules from the relationship between the visual attributes and the target classes, and (3) a multi-criteria data selection module to select a small set of informative and diverse automatically labeled data for users to inspect and fix. 
{\tool} works iteratively. It starts with a small initial set of training images created by human users.
In each iteration, {\tool} first extracts the visual attributes of each labeled image. 
Then, it takes as input the visual attributes and the corresponding class label of the images to infer a set of labeling rules for each class. 
The inferred rules are then applied to automatically generate labels for unannotated images. In case of contradicting labels, {\tool} selects the corresponding label of the rule with the highest Clause Satisfaction Ratio (detailed in Section~\ref{sec:informativeness}, Equation~\ref{equ:csr}).
Next, {\tool} adopts a multi-criteria data selection strategy to compute the informativeness score for each unannotated data and select a diverse set of data samples for users to inspect.
Users can fix the incorrect labels, which will be used to refine the labeling rules by {\tool}.

\subsection{Visual Attribute Extraction with Pre-trained Models}
A visual attribute extractor processes a given image and extracts the basic, low-level attributes of the image that are useful and relevant to the labeling task. 
The visual attributes can be object types in an image, the relationships between the objects, and an object's properties (e.g., size, number). This work mainly uses pre-trained perception models as the visual attribute extractors, detailed in Section~\ref{sec:dataset}. But one can also use a traditional feature extractor such as SIFT~\cite{790410} to extract visual attributes.
The visual attribute extractors are designed as pluggable components in our approach. For different labeling tasks, we use different pre-trained models to extract visual attributes related to the labeling task. This design increases the flexibility of our approach to be reused for new labeling tasks. 

\subsection{Rule Inference via Inductive Logic Learning}
\label{sec:foil}
Given its capability to learn from a small amount of data, we use First Order Inductive Learner (FOIL)~\cite{quinlan1990learning} to infer labeling rules. Furthermore, the declarative nature of logic rules makes them easy to be understood and refined by human labelers based on their domain knowledge. FOIL is initially designed to learn a logic rule with pre-defined predicates to distinguish a set of positive and negative examples. In our design, each predicate represents one trait of a visual attribute. 
The original FOIL algorithm can only infer logic rules with variables, which lacks the expressiveness for logic rules with constant values. Therefore, we extend FOIL to support the inference of constant values. As a consequence, this increases the search space exponentially. To address this challenge, we design several inductive biases, such as a TF-IDF-based heuristic, to improve search efficiency (detailed in Appendix~\ref{sec_supp:FOIL}).
\begin{table}[h]
\centering
\caption{Logic Predicates for Expressing Visual Attributes}
\label{tab:predicates}
\resizebox{0.8\columnwidth}{!}{%
\begin{tabular}{@{}ll@{}}
\toprule
\textbf{Predicate} & \textbf{Description} \\ \midrule
object\texttt{(X,A)} & Object \texttt{A} exists in image \texttt{X} \\
overlap\texttt{(A,B)} & Object \texttt{A} and \texttt{B} overlap the image \\
color\texttt{(A,Y)} & The color of object \texttt{A} is \texttt{Y} \\
num\texttt{(A,N)} & There are \texttt{N} object \texttt{A} in the image \\
area\texttt{(A,N)} & Object \texttt{A} has the area of \texttt{N} in the image \\
greater\texttt{(N,$\alpha$)} & \texttt{N} is greater than \texttt{$\alpha$} \\
smaller\texttt{(N,$\alpha$)} & \texttt{N} is smaller than \texttt{$\alpha$} \\ \bottomrule
\end{tabular}
}
\end{table}

In our approach, a labeling rule is defined in a disjunctive normal form with $k$ clauses, as shown below.
\begin{equation}
L\gets C_1\vee C_2\vee...\vee C_m
\end{equation}
$C_1,...C_m$ denote clauses and $L$ denotes the label. If at least $1$ clause is satisfied, an image is labeled as class $L$.
A clause is defined as,
\begin{equation}
C\gets p_1\wedge p_2\wedge...\wedge p_k
\end{equation}
where $p_1,...p_k$ are logic predicates for visual attributes.
A clause is a conjunctive normal form with $k$ predicates. 
Hence, a clause is satisfied if and only if all the predicates are satisfied.
In this work, we design a set of primitive predicates for different kinds of visual attributes, as shown in  Table~\ref{tab:predicates}. 
For example, in the traffic scenario labeling task, a target image class, ``highway'', can be inferred based on the types of objects on the road, e.g., ``trucks''. An example labeling rule for ``highway'' images can be:
\begin{equation}
\label{equ:rule_highway}
\begin{split}
high&way(\texttt{X})\gets \neg people\texttt{(X,B)} \vee \\
    &(truck\texttt{(X,A)}\wedge num\texttt{(A,N)}\wedge greater\texttt{(N,5)})
\end{split}
\end{equation}
where \texttt{X} is an input image, \texttt{A} and \texttt{B} are objects detected by a pre-trained object detection model. This rule means that if there are no pedestrians or there exist more than five trucks, the image is classified as ``highway''. In practice, users can redesign the predicates, e.g., by removing irrelevant predicates and adding domain-specific predicates based on the characteristics of each labeling task to improve the efficiency of inferring logic rules.

\begin{algorithm}[ht]
\renewcommand{\algorithmicrequire}{\textbf{Input:}}
\renewcommand{\algorithmicensure}{\textbf{Output:}}
\caption{Inductive Learning for Labeling Rules}
\label{alg:FOIL}
\begin{algorithmic}[1]
\REQUIRE {positive examples ($T^{+}$) and negative examples ($T^{-}$) for a label, clauses that must be included~($I$) and must be excluded~($E$)}
\ENSURE {rule R}
\STATE $R \leftarrow$ $\varnothing$
\STATE $R$.\textsc{append}($I$)
\WHILE {$T^{+}$ != $\varnothing$}
\STATE \textit{clause} $\leftarrow$ $\varnothing$
\STATE \textit{$T_i^{-}$}$\leftarrow$ \textit{$T^{-}$}
\STATE \textit{S} $\leftarrow$ \textsc{Initialize}(\textit{$T^{+}$})
\WHILE {$T_i^{-}$ != $\varnothing$}
\STATE \textit{clause}.\textsc{append}(\textsc{Max\_Gain}(\textit{S}))
\STATE \textit{$T_i^{-}$}$\leftarrow$ \textsc{remove}(\textit{$T^{-}$,clause})
\ENDWHILE
\IF{\textit{clause} $\notin$ $E$}
\STATE \textit{$T^{+}$}$\leftarrow$ \textsc{remove} (\textit{$T^{+}$,clause})
\STATE $R$.\textsc{append}(\textit{clause})
\ENDIF
\ENDWHILE
\RETURN $R$
\end{algorithmic}
\end{algorithm}

Algorithm~\ref{alg:FOIL} describes how to infer labeling rules with inductive learning. For each image label, our algorithm takes the set of positive examples \textit{$T^{+}$} and the set of negative examples \textit{$T^{-}$} as input and infers a logic rule. It also allows human labelers to specify which clauses must be included ($I$) or excluded ($E$) based on their domain knowledge.
It first adds must-include clauses into the rule $R$ (Line {2}).
Then, it keeps searching for possible clauses until \textit{$T^{+}$} is empty (Lines {3} to {15}). 
If the clause does not need to be excluded (Line {11}), the algorithm removes the positive examples which contain all visual attributes in the clause from $T^{+}$ (Line {12}), and then adds the clause to the rule~(Line~{13}). 
The algorithm initializes a negative set \textit{$T_i^-$} as \textit{$T^-$} ~(Line~{5}) and a set containing all possible predicates from the set of positive examples $S$~(Line~{6}) before the first iteration of the inner loop. 
To find a possible clause (Line {7} to {10}), the algorithm constantly selects predicates with the maximum information gain from $S$ and adds into the clause (Line {8}) until \textit{$T_i^{-}$} is empty (Line {7}). 
The information gain of each predicate is defined below:
\begin{equation}
\resizebox{0.75\hsize}{!}{$Gain(S_i)=T_i^{++} \times(log_2(\frac{T_{i+1}^+}{T_{i+1}^++T_{i+1}^-})-log_2(\frac{T_{i}^+}{T_{i}^++T_{i}^-}))$} \label{gain}
\end{equation}
where $T_i^+$ and $T_i^-$ denote the set of positive examples and the set of negative examples {\em before} adding the new predicate $S_i$.
$T_{i+1}^+$ and $T_{i+1}^-$ denote the set of positive examples and the set of negative examples {\em after} adding $S_i$. 
Then in each iteration in the inner loop, $T_i^-$ is redefined to a set that removes the negative examples which contain all visual attributes in the clause from $T^-$(Line {9}). The loop continues until it finds a labeling rule that matches all labeled images in a given class (i.e., positive examples) while not matching labeled images in other classes (i.e., negative examples).

\subsection{Labeling Rule Refinement via Active Learning}
\label{sec:active_learning}
Due to the ambiguity and incompleteness of the small amount of training data, the inductive learning module may not learn the best labeling rules in one pass. 
We propose to use active learning to improve the performance of inductive learning by iteratively soliciting more human labels.

In each iteration of active learning, when selecting data samples, our goal is to choose the data with the most information and variety to reduce data usage and improve performance. We propose a multi-criteria data selection strategy to achieve this goal.

\begin{algorithm}[ht]
\renewcommand{\algorithmicrequire}{\textbf{Input:}}
\renewcommand{\algorithmicensure}{\textbf{Output:}}
\caption{Multi-criteria Data Selection}
\label{alg:select}
\begin{algorithmic}[1] 
\REQUIRE {Unlabeled data $U$, the size of the intermediate set $M$, the number of data samples to select $N$}
\ENSURE {The set of selected data samples $S$}
\STATE Calculate the informative score for $U$
\STATE $U_{ranked}$ $\gets$ sort $U$ by informativeness score
\STATE $S_{intermediate}$ $\gets$ pick top $M$ data instances
\STATE $S$ $\gets$ \textsc{K-Means}($S_{intermediate}$)
\RETURN $S$
\end{algorithmic}
\end{algorithm}

\subsubsection{Multi-Criteria Data Selection}

Algorithm~\ref{alg:select} describes the multi-criteria data selection process.
First, it calculates the informativeness score of each unlabeled data instance and ranks them based on the scores. 
Then, it selects the $M$ most informative samples to form an intermediate set. These samples are then clustered based on similarity, and our algorithm selects the final set of $N$ samples that are both informative and diverse. The informativeness metric and the clustering algorithm are detailed in the following subsections.

\subsubsection{Informativeness}
\label{sec:informativeness}
Existing informativeness metrics in the literature of active learning are typically calculated based on prediction probabilities, e.g., entropy-based uncertainty~\cite{joshi2009multi}. Thus, they are only applicable to statistical models that output prediction probabilities. Since our approach uses logical rules, it does not produce a probability for an inferred image label. 

To bridge the gap, we propose a novel informativeness metric based on the extent of labeling conflicts among labeling rules. This design is based on the insight that multiple labeling rules may generate conflicting labels for the same image, which can be leveraged to measure the uncertainty of image labeling. The more conflicts there are in our labeling rules about the image labeling result of an image, the more information the image can bring to our model.  

The informativeness is largely measured by the number of inconsistent labels.
To break the tie among images with the same number of inconsistent labels,
we further consider the extent of conflict in the unsatisfied labels ($U$ in Equation~\ref{equ:score}).
\begin{equation}
\label{equ:score}
    Score(i)=
        \begin{cases}
            0,&\text{\#label}=1\\
            \lambda( \text{\#label} )+ U(i),&\text{Otherwise}
        \end{cases}
\end{equation}
\begin{equation}
\label{equ:utility}
    U(i)=1-\frac{\sum_{r \in \textsc{Unsatisfied}} CSR(i, r)}{\left| \textsc{Unsatisfied} \right|}
\end{equation}
%
In the equation above, $i$ denotes an image; $R$ denotes a labeling rule composed of a disjunction of clauses ($L\gets C_1 \vee C_2 \vee ... \vee C_m$). Here, each rule exclusively defines a unique label.
$\textsc{Unsatisfied}$ denotes the set of rules the image $i$ does not satisfy.
$\lambda$ is a hyper-parameter, which is empirically set to $0.6$ in our experiments.
The $CSR$---Clause Satisfactory Ratio---measures the degree of satisfaction of a single rule, as defined in Equation~\eqref{equ:csr}. 
$U$ measures the average $CSR$ for the rules that the image does not satisfy.
\begin{equation}
\label{equ:csr}
    CSR(i, r)=\max_{i=1..m}\frac{\sum_{j=1}^{k} Sat(i, p_{ij})}{k}
\end{equation}
In Equation~\eqref{equ:csr}, $p_{ij}$ denotes the $j$-th predicate of the $i$-th clause in rule \textit{R}. 
$Sat(i, p)$ represents whether a predicate is satisfied:
\begin{equation}
\label{equ:satl}
    Sat(i, p)=
        \begin{cases}
            1,&p \text{ is satisfied by } i\\
            0,&\text{Otherwise}
        \end{cases}
\end{equation}

\subsubsection{Diversity}
\label{sec:diversity}
The robustness of image labeling models largely depends on the variety of labeled data.
Therefore, our data selection algorithm also accounts for the diversity of images to maximize the variety of a group of data samples when selecting the data to label.  
We propose to cluster all the unlabeled images into $k$ clusters and choose one sample from each cluster to form the group of data to label.
We use k-Means to cluster the data samples and choose the final centroids to generate a diverse group of samples. 
In k-Means, each image is represented with a vector, and the similarity between every two images is measured by the cosine similarity of the two vectors.
The feature vector for an image has $d$ dimensions, which represent the total $d$ types of objects we consider for this task (or dataset). 
The value in each dimension is the number of the corresponding objects detected in the image.

%% file: 4-Experiments.tex
\section{Experiments}

We design experiments to answer the following research questions.
\begin{enumerate}
    \item[RQ1] How effective is {\tool} on image labeling tasks from different domains?
    \item[RQ2] How effective are the two kinds of human feedback solicited by {\tool}?
    \item[RQ3] How effective is inductive logic learning compared to statistical and neural network models?
    \item[RQ4] How sensitive is our active learning algorithm to different data selection strategies?
\end{enumerate}

\subsection{Experiment Design \& Setup}
To answer RQ1, we measure the image labeling accuracy and efficiency of {\tool} on four different image labeling tasks---two from highly specialized domains and two from common domains. Section~\ref{sec:dataset} describes the four tasks and datasets. To represent the condition of learning from limited training data, for each task, we bootstrap {\tool} with only $3$ randomly sampled data instances with labels to learn the initial set of rules. In each following iteration, the active learning module selects $3$ images and corrects their labels if wrong to refine the learned labeling rules. The choice of $3$ is to simulate the rapid, incremental feedback from users. This process continues until the $20$th iteration. Thus, for each task, {\tool} is trained with in total $60$ images. We compare {\tool} with $6$ baselines. Section~\ref{sec:baseline} describes these baselines and their training procedures. 


To answer RQ2, we measure the degradation of image labeling accuracy and efficiency when abating the human feedback mechanisms in {\tool}. {\tool} supports two kinds of human feedback---(1) directly editing the labeling rules generated by {\tool} and (2) fixing incorrect labels inferred by {\tool} and supplementing new labels. Thus, we create three variants of {\tool}---{\tool} without rule editing ({\tool$_{no\text{-}edit}$}),  {\tool} without any labeling correction ({\tool$_{no\text{-}al}$}), and {\tool} without any kinds of feedback ({\tool$_{no\text{-}feedback}$}).

RQ3 aims to measure the effectiveness of adopting inductive logic learning for image labeling. To answer RQ3, we create four variants of {\tool} by replacing the inductive logic learning module in {\tool} with three statistical models---SVM, random forest, and XGBoost~\cite{Chen:2016:XST:2939672.2939785}---and a neural network.
The design of these variants is inspired by existing frameworks such as Snorkel~\cite{ratner2017snorkel} and Concept Bottle Network~\cite{koh2020concept}, which use statistical models to make the final prediction based on symbolic representations extracted from raw input data. For the neural network variant, we adopt the design of the fully connected layers in ResNet-18~\cite{he2016deep}.

To answer RQ4, we compare {\tool} with three alternative data selection strategies---random selection, selection with only the informativeness criterion, and selection with only the diversity criterion.
We use image labeling accuracy, as well as the hit rate of misclassified data, as the evaluation metrics. Specifically, 
the hit rate is defined as the percentage of selected data samples that {\tool} mislabels in the current iteration and thus is worth fixing. A higher hit rate indicates better effectiveness of data selection.


\subsection{User Simulation}
Since {\tool} is designed as a human-in-the-loop approach, {\tool} needs to keep soliciting feedback from human experts to refine the labeling rules. It is expensive to recruit human participants to provide feedback, especially in the two highly specialized labeling tasks that require domain experts such as ophthalmologists and ornithologists. Therefore, we develop an automated script to simulate human feedback based on the ground truth data. To simulate label corrections, our script compares the ground truth label of each image with the labels inferred by {\tool} and automatically fixes the incorrect labels. To simulate human edits to labeling rules, the authors first manually constructed a set of high-quality labeling rules based on their own knowledge and the information shared on professional websites. In each iteration of the training process, our script compares the labeling rule inferred by {\tool} with the corresponding manually curated rule. Our script then replaces the first inconsistent clause with the clause in the manually curated rule. In all experiments, we restrict the simulation script to only edit one clause per iteration to simulate the incremental editing process of human labelers.

\subsection{Imagle Labeling Tasks and Datasets}
\label{sec:dataset}
{\tool} is designed for image labeling tasks in highly specialized domains.
Therefore, we first select two datasets---Glaucoma Diagnosis and Bird Species Labeling---from highly specialized domains.
To test the generalizability of {\tool}, 
we construct the two datasets on general domains, including traffic scene labeling and occupation labeling, by searching on Google and Flickr.

\textbf{Glaucoma Diagnosis.}
Given a color fundus image, this task requires labeling the eye in the image to be either normal or diseased.
We combine the color fundus images from three datasets, Drishti-GS~\cite{Drishti}, RIM-ONE\_r3~\cite{RIMONEDL}, and REFUGE~\cite{DBLP:journals/mia/OrlandoFBKBDFHK20}.  
We have $116$ images of glaucomatous eyes and $189$ images of normal eyes with both glaucoma diagnosis and structure segmentation.
We use a pre-trained model called BEAL~\cite{wang2019boundary} as the visual attribute extractor to obtain the segmentation of eye fundus structures
in the images. The visual attributes designed for this task are the diameter, area, and cup-to-disk ratio calculated based on the segmentation results.

\textbf{Bird Species Labeling.}
Given a bird image, this task requires labeling the bird species.
We use the Caltech-UCSD Birds-200-2011 (CUB 200-2011) dataset~\cite{WahCUB_200_2011}, containing $11,788$ bird images annotated with $200$ bird species and $312$ attributes that describe each body part of a bird, e.g., wing color, tail shape, etc. 
Following the experiment settings in Koh et al.~\cite{koh2020concept}, we use $112$ out of the $312$ attributes, and randomly choose three bird species to label in our experiments. 
We use the pre-trained concept models from Koh et al.~\cite{koh2020concept} to extract the visual attributes. 

\textbf{Occupation Labeling.}
Given an image of a person, this task requires labeling the occupation of the person.
For this task, we build a dataset containing $300$ images of three occupations---chef, farmer, and teacher. Each occupation has $100$ labeled images.
We use a pre-trained object detection model~\cite{Anderson2017up-down} to detect objects in the images and use the type (glasses, long hair, kitchen, etc.), color, and overlapping relationship between them as visual attributes.

\textbf{Traffic Scene Labeling.}
Given a road image, this task requires labeling the traffic scene of the image.
For this task, we build a dataset containing $420$ images of three traffic scenes---mountain road, highway, and downtown. Each traffic scene has $140$ labeled images. 
We use a pre-trained object detection model called DETR~\cite{detr} to detect the objects in the images and use the position, color, and type (e.g., pedestrian, truck, car, etc.) of the objects and the overlapping relationship between them as visual attributes.

\subsection{Comparison Baselines}
\label{sec:baseline}
For RQ1, we compare {\tool} with four image classification neural network baselines---ResNet-18~\cite{he2016deep}, ResNet-34, ResNeXt-32~\cite{Xie_2017_CVPR} and Inception-V3~\cite{szegedy2016rethinking})---and an active learning baseline called CEAL~\cite{DBLP:journals/corr/WangZLZL17}. We further compare {\tool} with GARDNet~\cite{mahrooqi2022gardnet} on the Glaucoma diagnosis task since GARDNet is specially designed for this task. 
To represent the condition of learning from limited training data, in each task, we randomly sample $30$ training data per class label to finetune the baseline models. That is a total of $60$ training samples for the Glaucoma Diagnosis task since it only has two class labels and $90$ for the other three tasks since they have three class labels.
For the first five baselines, we first pre-train them on ImageNet~\cite{5206848} and then fine-tune them on the four datasets, with training sets in the same size as training {\tool}. 
For GARDNet, we obtained the trained model from its original paper and then fine-tuned it on our Glaucoma diagnosis dataset.

For RQ2, we build three variants of {\tool}---{\tool} without editing rules by users ({\tool$_{no\text{-}edit}$}), {\tool} without labeling correction or new labels ({\tool$_{no\text{-}al}$}), and {\tool$_{no\text{-}feedback}$}, {\tool} without feedback when selecting training samples. Similar to the training setting of {\tool}, both {\tool$_{no\text{-}edit}$} and {\tool$_{no\text{-}al}$} are initially trained with $3$ randomly selected images. In the following interactions, {\tool$_{no\text{-}edit}$} only fixes incorrect labels in the $3$ images selected by the multi-criteria active learning algorithm per iteration but does not apply any direct edits to the inferred rules. By contrast, {\tool$_{no\text{-}al}$} only makes direct edits to one clause of the inferred rules per iteration but does not fix any incorrect labels. 
{\tool$_{no\text{-}feedback}$} does not use any active feedback from users. It is trained with randomly sampled images in various numbers (e.g., 3, 6, 9, etc.) ahead of time without soliciting further human feedback.

For RQ3, we create four variants of {\tool} by replacing the inductive logic learner with other machine learning methods, including SVM, random forest, gradient boosting, and neural network.
For these variants, we use a feature vector generated with the extracted visual attributes to represent each image. 
The feature vector is constructed in the same way as calculating the diversity criterion in Section~\ref{sec:diversity}. We repeat each training three times and compute the average performance of each baseline. 

\section{Results}
\subsection{RQ1. Effectiveness on Different Labeling Tasks}
Table~\ref{tab:results} shows the image labeling accuracy of {\tool} in the four labeling tasks from different domains in comparison to the fine-tuned models. {\tool} outperforms all baselines on the two highly specialized domains (Glaucoma Diagnosis and Bird Species Labeling) by $11.75\%$ to $12.03\%$. Specifically, {\tool} achieves a high accuracy of $85.52\%$ and $86.11\%$ in these two tasks, respectively. The result shows {\tool} can effectively infer accurate labeling rules in highly specialized domains with a small amount of training data.
For more details about the labeling rules, such as examples and statistics of the optimal rules, and how the labeling rules change over iterations, please refer to Appendix~\ref{sec_supp:rules}.

\begin{table}[ht]
\caption{Comparison of accuracy (\%) between {\tool} and image labeling baseline models in the four tasks.}
\label{tab:results}
\resizebox{0.85\columnwidth}{!}{%
\begin{tabular}{@{}lcccc@{}}
\toprule
 & \multicolumn{2}{c}{Highly Specialized Domains} & \multicolumn{2}{c}{Common Domains} \\ \cmidrule(l){2-5} 
 & Glaucoma & Bird Species & Occupation & Traffic \\ \midrule
ResNet-18 & 62.84 & 74.08 & 87.78 & 63.09 \\
ResNet-34 & 72.13 & 65.74 & \textbf{97.78} & 74.21 \\
ResNext-32 & 55.74 & 57.41 & 93.33 & 54.37 \\
Inception-V3 & 50.82 & 58.33 & 94.44 & 54.76 \\
CEAL & 73.77 & 66.67 & 89.99 & \textbf{92.86} \\
GARDNet & 54.65 & - & - & - \\
{\tool} & \textbf{85.52} & \textbf{86.11} & 88.33 & 83.33 \\ \bottomrule
\end{tabular}%
}
\end{table}

For the two common domains, {\tool} achieves comparable or worse accuracy. Specifically, {\tool} achieves an accuracy of $83.33\%$ in the traffic scene labeling task, while the best baseline model achieves $92.86\%$ accuracy. For the occupation labeling task, the accuracy of {\tool} is $88.33\%$ while the best baseline model achieves $97.78\%$ accuracy. This result is not surprising since the baseline models are pre-trained on ImageNet, which includes a considerable number of images similar to the ones in these two common tasks. Thus, the baseline models have already learned from many similar cases during the pre-training process. 

\begin{figure}[ht]
    \centering
    \includegraphics[width=0.95\linewidth]{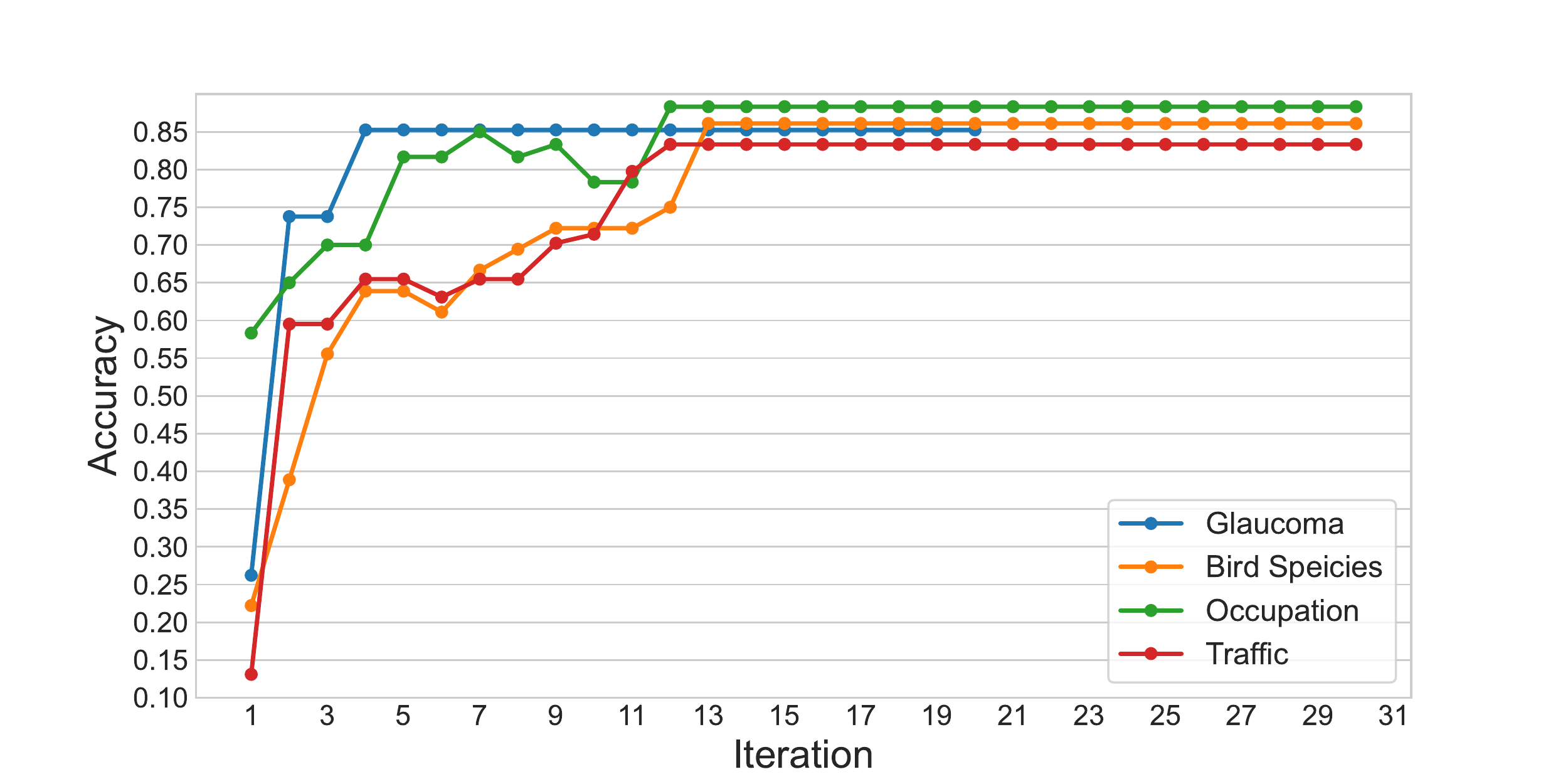}
    \caption{Image labeling accuracy of {\tool} on four tasks during the training process.}
    \label{fig:results}
\end{figure}

Figure~\ref{fig:results} shows the image labeling accuracy of {\tool} during the training process.
At the $4$th iteration, with only $12$ training samples, {\tool} has already achieved a reasonable accuracy---$85\%$ in  Glaucoma diagnosis, $70\%$ in occupation labeling, $65\%$ in traffic scene labeling, and $64\%$ in bird species labeling. Within the $13$th iteration, {\tool} has achieved the peak accuracy on all four tasks. 
Besides, during the training process, the performance of {\tool} is stably improving. The result shows {\tool} can effectively learn and refine labeling rules within a small number of iterations.

\subsection{RQ2. Effectiveness of Human Feedback}
Table~\ref{tab:feedback} shows the image labeling accuracy and the number of iterations to achieve the optimal accuracy of {\tool} in comparison to its invariants after ablating each feedback mechanism. Overall, {\tool} always achieves the highest accuracy with the smallest number of iterations ($10.25$ on average). {\tool}$_{no\text{-}feedback}$ has the worst performance with the largest number of iterations ($25.25$ on average). Without direct rule editing, {\tool}$_{no\text{-}edit}$ takes significantly 4X more iterations to achieve a comparable accuracy on the Glaucoma diagnosis task and has significantly lower accuracy on the other three tasks ($11.87\%$ accuracy decrease on average).

\begin{table}[htb]
\caption{Comparison of accuracy (\%) and the number of iterations consumed (Iter.) between {\tool} and three variants ablating two types of user feedback in the four tasks.}
\label{tab:feedback}
\resizebox{\columnwidth}{!}{%
\begin{tabular}{@{}lcrcrcrcr@{}}
\toprule
\multicolumn{1}{l}{} & \multicolumn{4}{c}{Highly Specialized Domains} & \multicolumn{4}{c}{Common Domains} \\ \cmidrule(l){2-9} 
\multicolumn{1}{l}{} & \multicolumn{2}{c}{Glaucoma} & \multicolumn{2}{c}{Bird Species} & \multicolumn{2}{c}{Occupation} & \multicolumn{2}{c}{Traffic} \\ \cmidrule(l){2-9} 
\multicolumn{1}{l}{} & Acc. & \multicolumn{1}{c}{Iter.} & Acc. & \multicolumn{1}{c}{Iter.} & Acc. & \multicolumn{1}{c}{Iter.} & Acc. & \multicolumn{1}{c}{Iter.} \\ \midrule
{\tool}$_{no\text{-}edit}$ & \textbf{85.25} & 21 & 66.67 & 13 & 81.67 & 27 & 73.81 & 10 \\
{\tool}$_{no\text{-}al}$ & \textbf{85.25} & 4 & \textbf{86.11} & 19 & \textbf{88.33} & 11 & \textbf{83.33} & 8 \\
{\tool}$_{no\text{-}feedback}$ & 83.61 & 14 & 55.56 & 28 & 80.00 & 29 & 67.86 & 30 \\
{\tool} & \textbf{85.25} & 4 & \textbf{86.11} & 13 & \textbf{88.33} & 12 & \textbf{83.33} & 12 \\ \bottomrule
\end{tabular}%
}
\end{table}

Though {\tool}$_{no\text{-}al}$ can achieve the same final accuracy as {\tool}, it takes $6$ more iterations in the bird species labeling task. It is interesting to observe that {\tool}$_{no\text{-}al}$ takes the same or even fewer iterations in three tasks. This is because the user simulation script always makes the right edit based on the ground-truth labeling rule each time. When using active learning together with rule editing, {\tool} will regenerate the label after receiving new labels in each iteration. These newly generated rules may deviate from the ground-truth rules, therefore leading to more iterations. 



Compared with the three variants, {\tool} achieves the largest performance gain in the bird species labeling task.
This performance gain can be largely attributed to the inherent learning challenge of the dataset. For example, 
birds of the same species are of great variety (e.g., a Kentucky Warbler can exhibit eight distinct wing color variations.). Thus, this increases the difficulty of learning a proper labeling rule to define what a certain bird species look like from such a small training dataset. 
On the other hand, 
by editing the rules, experts can directly embed their expert knowledge into the labeling rules, leading to a huge performance improvement. 

\begin{figure}[h]
    \centering
    \includegraphics[width=0.95\linewidth]{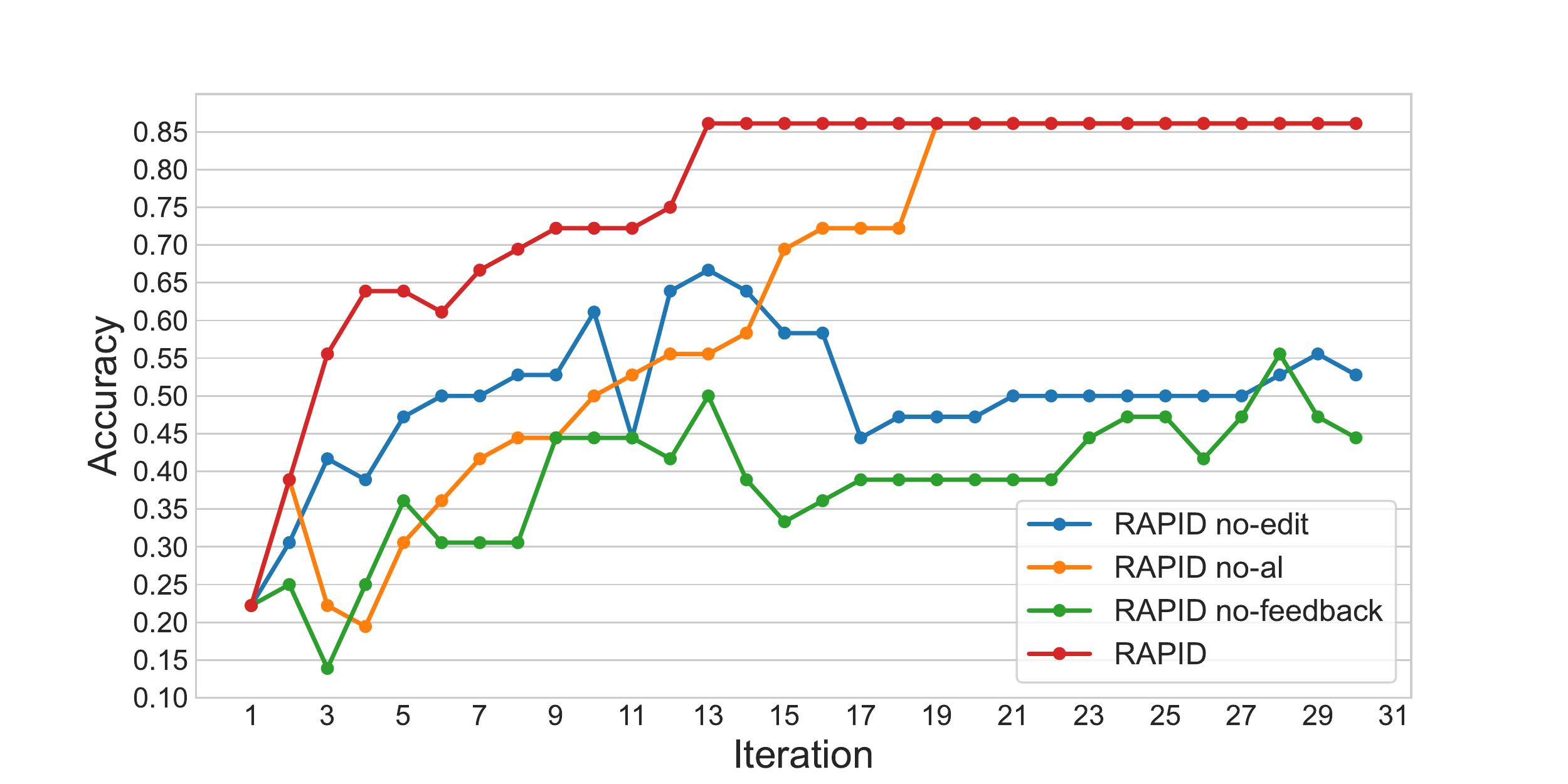}
    \caption{Comparison of accuracy in the training process between {\tool} and three variants ablating two types of user feedback in the bird species labeling task.} 
    \label{fig:edit_iter_bird}
\end{figure}
Figure~\ref{fig:edit_iter_bird} shows the image labeling accuracy of {\tool} and its variants over iterations in the bird species labeling task. 
For ease of comparison, we show the accuracy of {\tool}$_{no\text{-}feedback}$ over the number of randomly sampled training samples it uses. In other words, for {\tool}$_{no\text{-}feedback}$ in Figure~\ref{fig:edit_iter_bird}, $1$ in the x-axis means training with $3$ samples, $2$ means training with $6$ samples, $3$  means training with $9$ samples, so on and so forth. 
Overall, {\tool} achieves the highest image labeling accuracy ($86.11\%$) with only $12$ iterations and $36$ images in total. While 
{\tool}$_{no\text{-}al}$ achieves the same accuracy, it takes $7$ more iterations and thus $21$ more images to reach this accuracy. In this task, both {\tool}$_{no\text{-}edit}$ and {\tool}$_{no\text{-}feedback}$ takes significantly more iterations to achieve the final accuracy. 

\begin{table}[th]
\caption{Comparison of accuracy ($\%$) between {\tool} and four variants replacing the inductive logic learner with statistic and NN modules in the four tasks.}
\label{tab:replace}
\resizebox{0.95\columnwidth}{!}{%
\begin{tabular}{@{}lcccc@{}}
\toprule
 & \multicolumn{2}{c}{Highly Specialized Domains} & \multicolumn{2}{c}{Common Domains} \\ \cmidrule(l){2-5} 
 & Glaucoma & Bird Species & Occupation & Traffic \\ \midrule
{\tool}$_{SVM}$ & 50.82 & 70.37 & 41.11 & 33.33 \\
{\tool}$_{XGBoost}$ & 79.23 & 58.33 & 78.33 & 74.60 \\
{\tool}$_{RandomForest}$ & 79.23 & 70.37 & 79.44 & 79.76 \\
{\tool}$_{NeuralNetwork}$ & 41.53 & 42.59 & 33.55 & 34.52 \\
{\tool} & \textbf{85.25} & \textbf{86.11} & \textbf{88.33} & \textbf{83.33} \\ \bottomrule
\end{tabular}%
}
\end{table}

\subsection{RQ3. Effectiveness of Inductive Logic Learning}
Table~\ref{tab:replace} shows the comparison of accuracy between {\tool} and variants replacing the inductive logic learner with statistical and neural network models.
{\tool} outperforms all variant models in all four tasks by $6.02\%$, $15.74\%$, $8.89\%$ and $3.57\%$, respectively.
The results show the great capability of the inductive logic learning model in learning image labeling tasks under a low resource setting.

Though {\tool$_{RandomForst}$} does not achieve an accuracy as high as {\tool}, it is worth mentioning that {\tool$_{RandomForst}$} outperforms all the baseline models in Table~\ref{tab:results}, in the two highly specialized domain tasks. 
This implies that our pipeline, similar to Concept Bottleneck Models~\cite{koh2020concept}, which disentangles the perception process (Visual Attribute Extraction) and the learning process, has great capability in highly specialized image labeling tasks.

\begin{figure}[bht]
\begin{subfigure}[b]{0.95\linewidth}
         \centering
         \includegraphics[width=\linewidth]{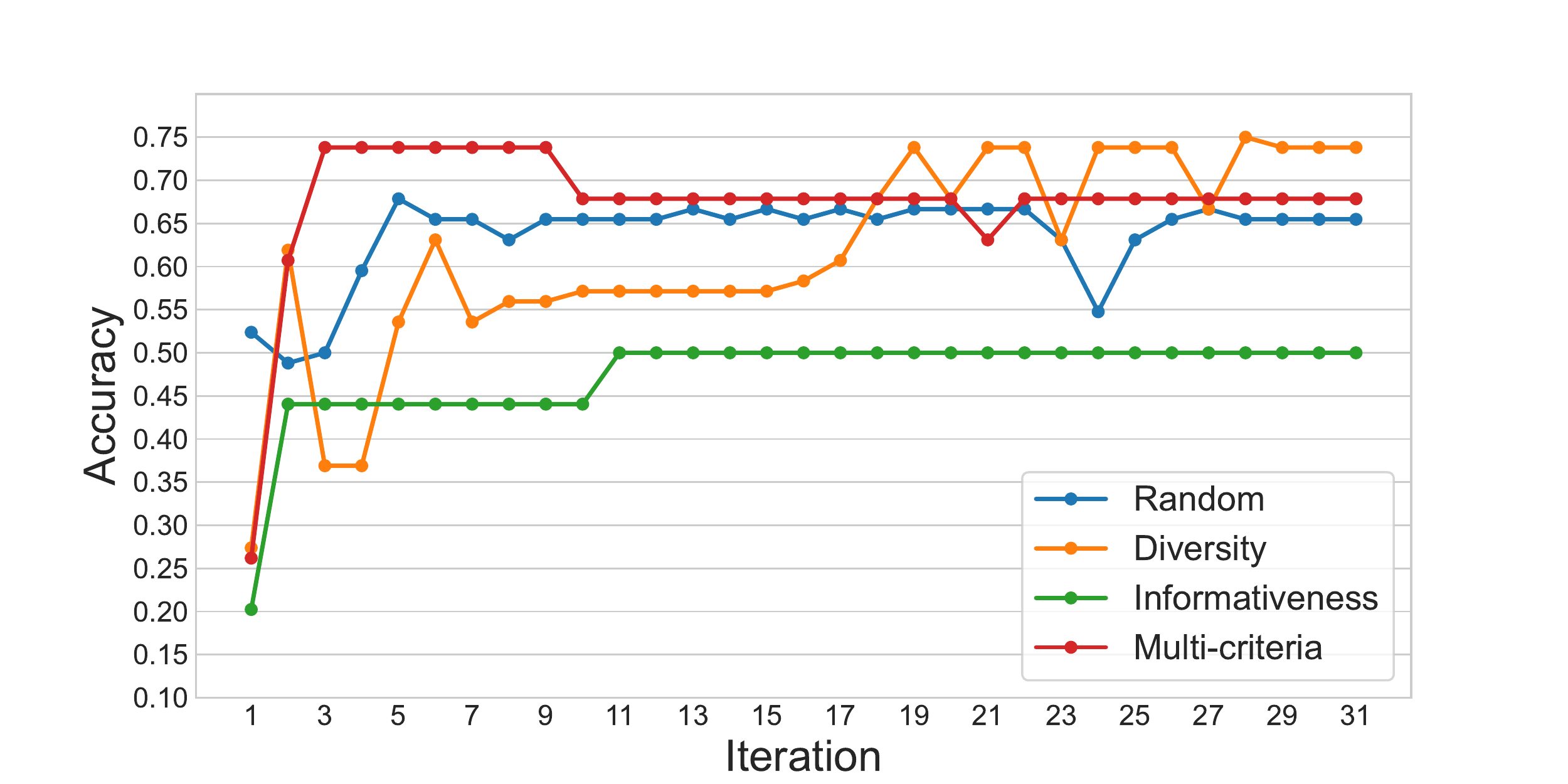}
         \caption{Comparison of accuracy.}
         \label{fig:selection_acc}
\end{subfigure}
     \hfill
\begin{subfigure}[b]{0.95\linewidth}
         \centering
         \includegraphics[width=\linewidth]{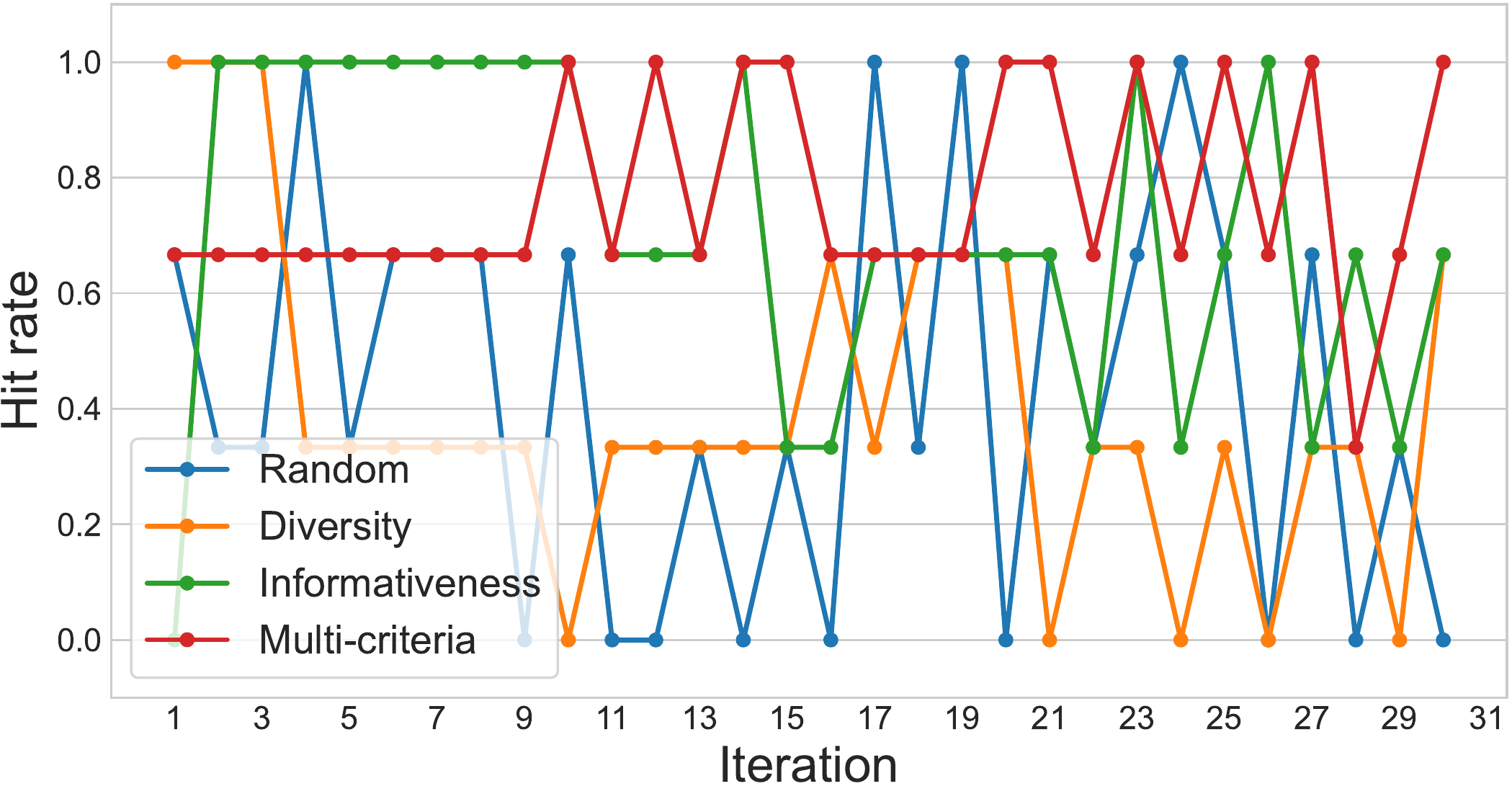}
         \caption{Comparison of hit rate.}
         \label{fig:selection_hit}
     \end{subfigure}
 \caption{Comparison of accuracy and hit rate for different data selection strategies in the traffic scene labeling task.}
 \label{fig:selection}
\end{figure}
\subsection{RQ4. Sensitivity Analysis of Active Learning}
Figure~\ref{fig:selection} shows the comparison of accuracy and hit rate between four data selection strategies in active learning in the traffic scene labeling task. Recall that the hit rate is defined as the percentage of selected data samples that {\tool} mislabels in the current iteration and thus is worth fixing. A higher hit rate indicates better effectiveness of data selection. 
Due to the page limit, we only showcase the experiment results on this task. Figures for the other three tasks have been provided in Appendix~\ref{sec_supp:rq4}.

Among the four data selection strategies, the multi-criteria strategy achieves the highest image labeling accuracy ($73.81\%$) with the smallest number of iterations ($3$). By contrast, using the diversity criterion alone takes $19$ iterations ($6$X more) to achieve similar accuracy. Both using the informativeness criterion alone and random selection achieve lower accuracy, $50.00\%$ and $67.86\%$. Yet compared with random selection, the informative selection still helps, achieving significantly higher accuracy within fewer iterations.

In the traffic scene labeling task, random selection, single diversity criterion, single informativeness criterion, and multi-criteria achieve $42.22\%$, $40.00\%$, $71.11\%$, and $76.67\%$ average hit rate over the iterations.
Combining informativeness and diversity, our multi-criteria strategy achieves the highest hit rate in this task.
The single informativeness strategy maintains the highest hit rate at the beginning of the training process, which proves the effectiveness of our informativeness metric designed in Section~\ref{sec:informativeness}.
The results on accuracy and hit rate imply that the two metrics are complementary to each other.
With the informativeness metric selecting training samples with high information and the diversity metric choosing representatives from the selected samples, {\tool} can achieve high accuracy in a small number of iterations.

%% file: 5-Discussion.tex
\section{Discussion}
The experiment results demonstrate the effectiveness of {\tool} in learning accurate labeling rules with a small amount of training data. Based on the ablation studies in RQ3 and RQ4, we found that both inductive logic learning and  multi-criteria active learning play vital roles in the success of {\tool}. Furthermore, given the inherent transparency and explainability of logic-based labeling rules, {\tool} provides affordance for users to directly embed their expertise into the labeling rules via rule editing. This further improves the efficiency of rule inference, leading to significantly fewer feedback iterations compared with using active learning alone. 

Our work is the first to apply inductive logic learning to neuro-symbolic learning. The experiment for RQ3 demonstrates the learning capability of our FOIL-based inductive logic learner. Specifically, our logic learner outperforms alternative statistical and neural network models. 
It is interesting to observe that when replacing the inductive logic learner with Random Forest, the resulting model can still achieve better performance than the fine-tuned models in Table~\ref{tab:results} in highly specialized domains. This implies that simply disentangling the perception process from the reasoning process can still be beneficial when learning highly specialized tasks in a low-resource setting (i.e., training with a small amount of data).  

Compared with Snorkel~\cite{ratner2017snorkel}, our approach has two significant advantages. First, {\tool} can learn an initial set of labeling rules from a small amount of labeled data (e.g., $3$ training samples in our case) as a starting point for expert users to refine. In contrast, Snorkel requires users to manually write labeling functions from scratch, which can be effortful for expressing complex knowledge.
Second, to use Snorkel, users need to be familiar with a programming language such as Python to write labeling functions. Thus, it comes with a steep learning curve for domain experts and end-users who do not have programming background. By contrast, the logic labeling rules in our approach are readable and intuitive. Thus, our approach comes with a more gentle learning curve compared with Snorkel.

This work is also a good demonstration of effective human-AI collaboration in challenging tasks. In our case, {\tool} infers an initial set of labeling rules and continuously refines them based on human feedback in the form of direct edits or label corrections. Our experiment in RQ2 has shown that incorporating human feedback is critical to infer accurate labeling rules. Without humans in the loop, {\tool} has significantly lower accuracy even when trained  with the same amount of data as training with many iterations.

Despite the promising results, the final image labeling accuracy of {\tool} may still not be on par with human experts. Our approach achieves an average of 85\% in the four labeling tasks, while the accuracy of human labelers on ImageNet is about 95\%~\cite{northcutt2021pervasive}. Similar to Snorkel~\cite{ratner2017snorkel}, we also want to argue that such relatively noisy data can still provide valuable supervision signals for model training, especially when used together with noise-robust learning~\cite{xue2019robust, wang2019symmetric} and weak supervision~\cite{xu2015learning, ratner2017snorkel}. This can be extremely beneficial in highly specialized domains such as medical imaging, where human labelers are expensive and hard to acquire.

The current design of the inductive logic learner in {\tool} has a rigid objective of learning a rule that matches all positive samples while rejecting all negative samples. Consequently, our approach is sensitive to user mistakes (e.g., incorrect labels and edits).  A single incorrect training sample can lead to a labeling rule that makes no sense to users. In future work, we will improve the inductive logic learning method by providing the flexibility to relax the rigid rule satisfaction requirement, which is expected to increase the noisy label tolerance and robustness.
Besides, when training {\tool} in this work, we follow the FOIL algorithm and use an information gain-based search method. In future work, We will explore improving the search mechanism by adding heuristics using visual attributes.

The performance of {\tool} highly depends on the visual attributes extracted by pre-trained computer vision models.
For specialized domains, it is possible that no pre-trained computer vision model can extract meaningful visual attributes.
However, we think this situation will not occur very often. There are two reasons. First, numerous large computer vision models have been developed and made available online (e.g., HuggingFace, GitHub) these days. These models can recognize a wide range of objects, shapes, colors, and other basic visual attributes that generally apply to different domains, including many highly specialized domains. Second, for rare visual attributes, the pre-trained models can be substituted with conventional handcrafted computer vision techniques, such as SIFT. 
Our inductive logic learning component can still act on those low-level attributes and synthesize meaningful labeling rules. 

%% file: 6-Conclusion.tex
\section{Conclusion}
We present a rapid image labeling method based on neuro-symbolic learning.
The proposed method uses pre-trained neural network models to extract visual attributes and use first-order inductive learning to infer labeling rules based on the visual attributes.
This architecture disentangles perception from learning, enabling our method to be applied to new tasks by easily changing the pre-trained visual attribute extractor. Besides, the declarative nature of logic rules enables users to directly inspect and edit the inferred labeling rules, explicitly embedding human expertise into the rules.
The experiments show that our method can achieve outstanding performance on highly specialized domains under an extremely low resource setting while generalizing to other general domains with reasonable performance.

%% file: 7-Appendix.tex
\newpage
\section{Details of inductive logic learning}
\label{sec_supp:FOIL}
To support the inference of constant values in our algorithm and keep the high search efficiency, we design several inductive biases. 

\subsection{Predicate Preprocessing}
\label{supp:predicate}
For the predicate set \textit{S} in Algorithm~1
{\tool} does some preprocessing before searching for the predicate with max information gain. The algorithm calculates the significance of each predicate using equation~\eqref{significance}:
\begin{equation}
\label{PF}
PF(S)=\frac{\sum_{i=1}^{len(T^+)} \textsc{In}(T_i^+,S_i)}{len(T^+)}
\end{equation}
\begin{equation}
\label{ISF}
ISF(S)=ln(\frac{len(T)}{\sum_{i=1}^{len(T)} \textsc{In}(T_i,S)})
\end{equation}
\begin{equation}
\label{significance}
Sig(S)=PF(S)\times ISF(S)
\end{equation}
where \textit{T} is the set of examples and \textit{$T^+$} is the set of positive examples. \textit{len($T^+$)} means the number of positive examples in \textit{$T^+$}, and \textit{len(T)} means the number of examples in \textit{T}. \textsc{In}\textit{($\upsilon$,$S$)} is defined in ~\eqref{satisfy}:
\begin{equation}
\label{satisfy}
\resizebox{0.91\hsize}{!}{
$\textsc{In}(\upsilon,S)=\left\{
\begin{array}{rcl}
1       &      & {predicate\; in\; the\; example\; \upsilon}\\
0       &      & {predicate\; not\; in\; the\; example\; \upsilon}
\end{array} \right.$}
\end{equation}
We set a threshold $\Theta$ to define the significance. If \textit{Sig($S_i$)} is bigger than $\Theta$, then it means the predicate is significant enough for the rule generation and should be reserved. Otherwise, it will be treated as noise by the algorithm and be dropped.

\subsection{Predicate Preference}
By analyzing the predicates in each rule, we find that the predicates about object type appear much more frequently than any other predicate.
Therefore, we design an inductive bias that the algorithm prioritizes searching for those predicates about object type. 
Only when the number of predicates about object type in the clause exceeds the threshold $\Phi$ will the algorithm search for other possible predicates. 
Besides, we also set separate thresholds for separate predicates to generate a rule which can result in a higher labeling accuracy. 

\section{Details of labeling rules}
\label{sec_supp:rules}
\subsection{Statistics of the labeling rules on the four tasks}

\begin{table}[tbh]
\caption{Statistics of the labeling rules on the four tasks.}
\label{tab:stat_rules}
\resizebox{\columnwidth}{!}{%
\begin{tabular}{lccccc}
\hline
 & \begin{tabular}[c]{@{}c@{}}\# of \\ rules\end{tabular} & \begin{tabular}[c]{@{}c@{}}\# of \\ Clauses\end{tabular} & \begin{tabular}[c]{@{}c@{}}avg. \# of \\ Clauses per Rule\end{tabular} & \begin{tabular}[c]{@{}c@{}}\# of \\ Predicates\end{tabular} & \begin{tabular}[c]{@{}c@{}}avg. \# of \\ Predicate per Clause\end{tabular} \\ \hline
Glaucoma & 2 & 2 & 1.0 & 6 & 3.0 \\
Bird Species & 3 & 29 & 9.7 & 52 & 1.8 \\
Occupation & 3 & 10 & 3.3 & 15 & 1.5 \\
Traffic & 3 & 7 & 2.3 & 8 & 1.1 \\ \hline
\end{tabular}%
}
\end{table}

We list the number of rules and the total number of clauses in those rules for each dataset in Table~\ref{tab:stat_rules}.

\subsection{The optimal rules for each task}
\label{sec:app_opti_rule}
The optimal rules on the four tasks are shown in Table~\ref{tab:rule_optimal}.

\begin{table}[htbp]
\caption{The optimal rules on the four tasks.}
\label{tab:rule_optimal}
\resizebox{\columnwidth}{!}{%
\begin{tabular}{l|l}
\hline
Task & Optimal Rules \\ \hline
Glaucoma & \begin{tabular}[c]{@{}l@{}}\textbf{Normal}(X): [ACDR(X,A) $\wedge$  area(A,N) $\wedge$ smaller (N,0.31)], \\ \textbf{Glaucoma}(X): [ACDR(X,A) $\wedge$ area(A,N) $\wedge$ greater(N,0.31)]\end{tabular} \\ \hline
Bird Species & \begin{tabular}[c]{@{}l@{}}\textbf{Orange crowned Warbler}(X): \\ \text{[}$\neg$ has eye color(X,black) \\ $\vee$ ($\neg$ has forehead color(X,grey) $\wedge$ has underparts color(X,olive)) \\ $\vee$ ($\neg$ has leg color(X,black) $\wedge$ has eye color(X,black)) \\ $\vee$ has forehead color(X,green)\\ $\vee$ ($\neg$ has leg color(X,black) $\wedge$ has head pattern(X,plain))\\ $\vee$ (has nape color(X,yellow) $\wedge \neg$ has leg color(X,black))\\ $\vee$ (has under tail color(X,yellow) $\wedge$ has leg color(X,buff)) \\ $\vee$ (has nape color(X,buff) $\wedge \neg$ has under tail color(X,yellow))\\ $\vee$ has forehead color(X,olive) \\ $\vee$ ($\neg$ has nape color(X,grey) $\wedge \neg$ has under tail color(X,yellow) \\ ~~~~~~~~~~~~$\wedge$ has head pattern(X,eyeline)) \\ $\vee$ (has under tail color(X,brown) $\wedge$ has forehead color(X,brown))\\ $\vee$ (has nape color(X,yellow) $\wedge$ has forehead color(X,yellow) \\ ~~~~~~~~~~~~$\wedge$ has underparts color(X,yellow)) \\ $\vee$ (has forehead color(X,buff) $\wedge$ has bill color(X,buff) \\ ~~~~~~~~~~~~$\wedge$ has underparts color(X,buff)) \\ $\vee$ (has leg color(X,yellow) $\wedge$ has belly color(X,black)) \\ $\vee$ has breast color(X,red)], \\ \textbf{Nashville Warbler}(X): \\ \text{[}$\neg$ has under tail color(X,yellow) \\ $\vee$ (has belly color(X,yellow) $\wedge$ has forehead color(X,grey) \\ ~~~~~~~~~~~~$\wedge \neg$ has under tail color(X,yellow)) \\ $\vee$ has forehead color(X,grey) \\ $\vee$ (has belly color(X,yellow) $\wedge$ has nape color(X,buff)) \\ $\vee$ (has nape color(X,grey) $\wedge$ has primary color(X,yellow)) \\ $\vee$ has shape(X,sandpiper-like) \\ $\vee$ has leg color(X,buff)], \\ \textbf{Myrtle Warbler}(X): \\ \text{[}($\neg$ has belly color(X,yellow) $\wedge$ has eye color(X,black)) \\ $\vee \neg$ has nape color(X,grey) \\ $\vee$ (has leg color(X,black) $\wedge \neg$ has breast color(X,yellow)) \\ $\vee$ has breast color(X,white)\\ $\vee$ ($\neg$ has bill color(X,black) $\wedge$ has primary color(X,yellow)) \\ $\vee$ ($\neg$ has under tail color(X,yellow) $\wedge$ has breast color(X,buff)) \\ $\vee$ has bill color(X,black) \\ $\vee$ has underparts color(X,buff)]\end{tabular} \\ \hline
Occupation & \begin{tabular}[c]{@{}l@{}}\textbf{Cook}(X): \text{[}bowl(X,E) $\vee$ cabinet(X,B) $\vee$ food(X,F)], \\ \textbf{Teacher}(X): \\ \text{[}(wall(X,E)$\wedge$$\neg$counter(X,K) $\wedge$ $\neg$bowl(X,L)$\wedge$woman(X,A))\\ $\vee$ $\neg$man(X,B)\\ $\vee$ (wall(X,E)$\wedge$ $\neg$bowl(X,L))\\ $\vee$(color(C,brown)$\wedge$woman(X,A))], \\ \textbf{Farmer}(X): [tree(X,D) $\vee$$\neg$wall(X,C) $\vee$ grass(X,E)]]\end{tabular} \\ \hline
Traffic & \begin{tabular}[c]{@{}l@{}}\textbf{Downtown}(X):\text{[}building(X,A)],\\ \textbf{Highway}(X): \text{[}fence(X,F) \\ $\vee$ (¬mountain(X,B) $\wedge$ grass(X,A)) \\ $\vee$ $\neg$ mountain(X,B) \\ $\vee$ truck(X,G)],\\ \textbf{Mountain road}(X):\text{[}mountain(X,A) $\vee$ rock(X,C)]\end{tabular} \\ \hline
\end{tabular}%
}
\end{table}

\subsection{Change of rules along with iterations}
Two kinds of rule updates may occur in iterations of active learning.
First, one or more parameters in a rule may get updated with new labeled data samples provided by a user. Take the glaucoma diagnosis labeling task as an example. As shown in Table~\ref{tab:glaucoma_iter}, the parameters of the logic predicates, ``greater'' and ``smaller'', were updated in each iteration. 

Second, logic predicates or clauses may be added or removed in an iteration. Take the bird species labeling task as an example. As users provided more labeled images during the iterations, more logic predicates and clauses were added over the iterations to distinguish those examples. For simplicity, we show the labeling rule for one bird species—the orange-crowned warbler—at Iterations 1, 5, and 13 in table~\ref{tab:bird_iter}.

\begin{table}[htbp]
\caption{Change of rules along with iterations on the Glaucoma Diagnosis task and the corresponding accuracy}
\label{tab:glaucoma_iter}
\resizebox{\columnwidth}{!}{%
\begin{tabular}{@{}l|l|l@{}}
\toprule
Iter. & Labeling Rules & Acc. (\%) \\ \midrule
1 & \begin{tabular}[c]{@{}l@{}}\textbf{Normal}(X): [ACDR(X,A)$\wedge$ area(A,N)$\wedge$ smaller(N, 0.17)], \\ \textbf{Glaucoma}(X): [ACDR(X,A)$\wedge$ area(A,N)$\wedge$ greater(N, 0.62)]\end{tabular} & 26.23 \\ \midrule
2 & \begin{tabular}[c]{@{}l@{}}\textbf{Normal}(X): [ACDR(X,A)$\wedge$ area(A,N)$\wedge$ smaller(N, 0. 44)], \\ \textbf{Glaucoma}(X): [ACDR(X,A)$\wedge$ area(A,N)$\wedge$ greater(N, 0. 50)]\end{tabular} & 73.77 \\ \midrule
3 & \begin{tabular}[c]{@{}l@{}}\textbf{Normal}(X): [ACDR(X,A)$\wedge$ area(A,N)$\wedge$ smaller (N,0.24)], \\ \textbf{Glaucoma}(X): ACDR(X,A)$\wedge$ area(A,N)$\wedge$ greater(N,0.31)]\end{tabular} & 73.77 \\ \midrule
4 & \begin{tabular}[c]{@{}l@{}}\textbf{Normal}(X): [ACDR(X,A)$\wedge$ area(A,N)$\wedge$ smaller (N,0.31)]], \\ \textbf{Glaucoma}(X): [ACDR(X,A)$\wedge$ area(A,N)$\wedge$ greater(N,0.31)]\end{tabular} & 85.25 \\ \bottomrule
\end{tabular}%
}
\end{table}

\begin{table}[htbp]
\caption{Change of rules along with iterations on the Bird Species Labeling task (Orange-crowned Warbler) and the corresponding accuracy}
\label{tab:bird_iter}
\resizebox{\columnwidth}{!}{%
\begin{tabular}{@{}l|l|l@{}}
\toprule
Iter. & Labeling Rules & Acc. (\%) \\ \midrule
1 & \textbf{Orange crowned Warbler}(X): [has underparts color(X,yellow)] & 22.22 \\ \midrule
5 & \begin{tabular}[c]{@{}l@{}}\textbf{Orange crowned Warbler}(X): [$\neg$has eye color(X,black) \\ ~~~~$\vee$ has belly color(X,grey) \\ ~~~~$\vee$ has nape color(X,black) \\ ~~~~$\vee$ $\neg$has forehead color(X,grey)]\end{tabular} & 63.89 \\ \midrule
13 & \begin{tabular}[c]{@{}l@{}}\textbf{Orange crowned Warbler}(X): \text{[}$\neg$ has eye color(X,black) \\ $\vee$ ($\neg$ has forehead color(X,grey) $\wedge$ has underparts color(X,olive)) \\ $\vee$ ($\neg$ has leg color(X,black) $\wedge$ has eye color(X,black)) \\ $\vee$ has forehead color(X,green)\\ $\vee$ ($\neg$ has leg color(X,black) $\wedge$ has head pattern(X,plain))\\ $\vee$ (has nape color(X,yellow) $\wedge \neg$ has leg color(X,black))\\ $\vee$ (has under tail color(X,yellow) $\wedge$ has leg color(X,buff)) \\ $\vee$ (has nape color(X,buff) $\wedge \neg$ has under tail color(X,yellow))\\ $\vee$ has forehead color(X,olive) \\ $\vee$ ($\neg$ has nape color(X,grey) $\wedge \neg$ has under tail color(X,yellow) \\ ~~~~~~~~~~~~$\wedge$ has head pattern(X,eyeline)) \\ $\vee$ (has under tail color(X,brown) $\wedge$ has forehead color(X,brown))\\ $\vee$ (has nape color(X,yellow) $\wedge$ has forehead color(X,yellow) \\ ~~~~~~~~~~~~$\wedge$ has underparts color(X,yellow)) \\ $\vee$ (has forehead color(X,buff) $\wedge$ has bill color(X,buff) \\ ~~~~~~~~~~~~$\wedge$ has underparts color(X,buff)) \\ $\vee$ (has leg color(X,yellow) $\wedge$ has belly color(X,black)) \\ $\vee$ has breast color(X,red)]\end{tabular} & 86.11 \\ \bottomrule
\end{tabular}%
}
\end{table}

\section{Experimental Results of RQ4 in other three tasks.}
\label{sec_supp:rq4}
\begin{figure}[htbp]
\begin{subfigure}[b]{0.95\linewidth}
         \centering
         \includegraphics[width=\linewidth]{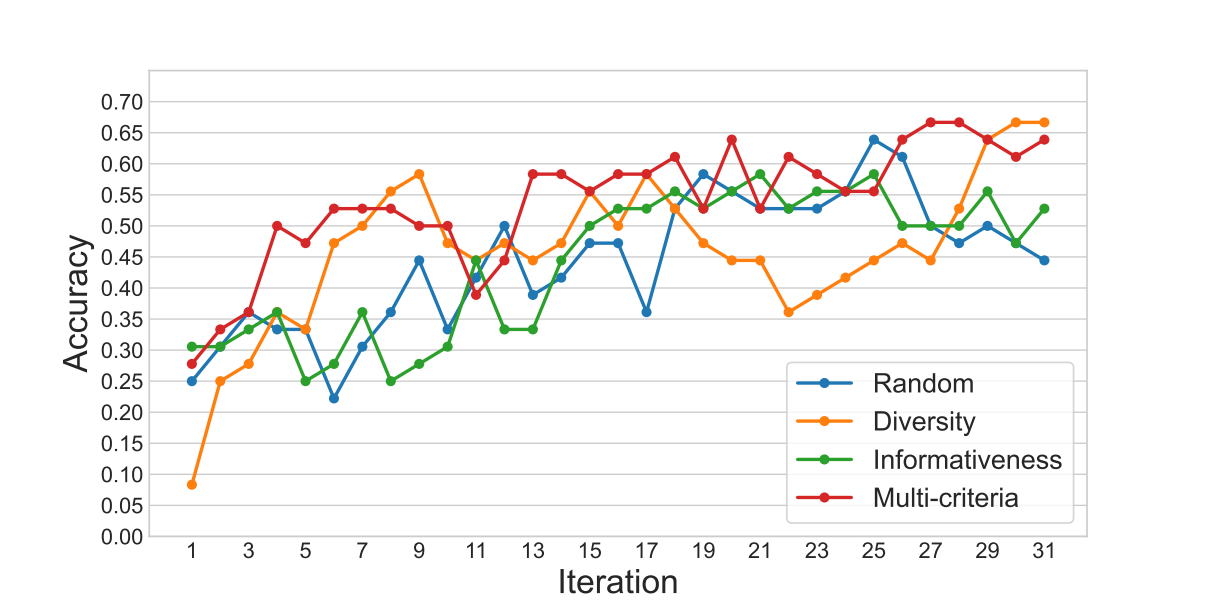}
         \caption{Comparison of accuracy.}
     \end{subfigure}
     \hfill
     \begin{subfigure}[b]{0.95\linewidth}
         \centering
         \includegraphics[width=\linewidth]{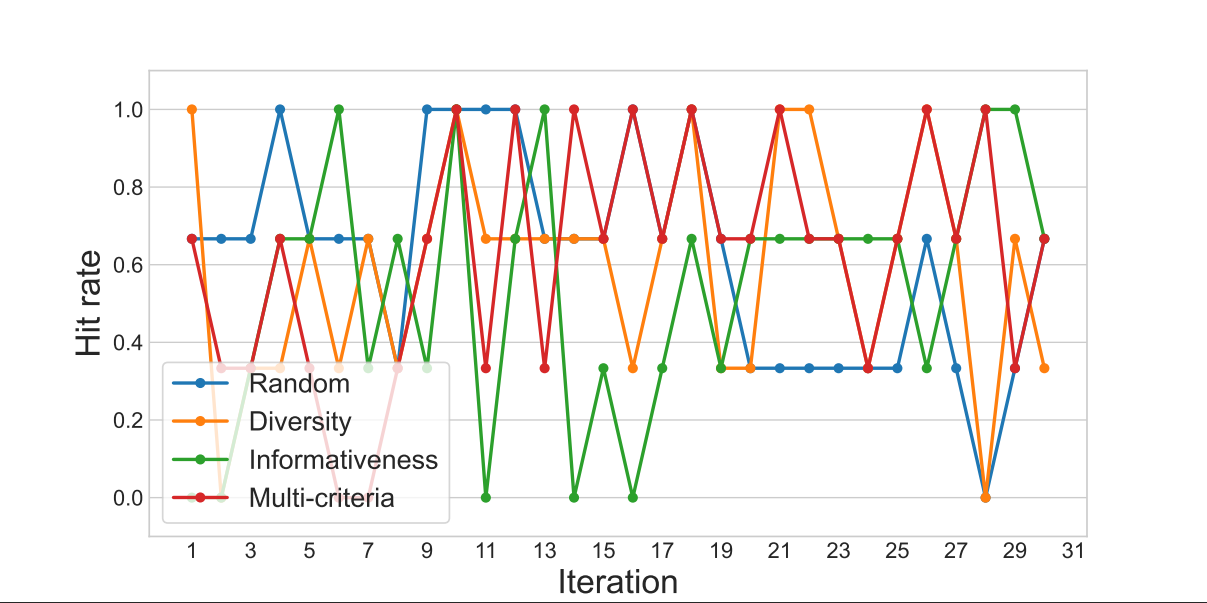}
         \caption{Comparison of hit rate.}
     \end{subfigure}
 \caption{Comparison of accuracy and hit rate in the bird species  labeling task for different data selection strategies.}
 \label{fig:selection_bird}
\end{figure}

\begin{figure}[htbp]
\begin{subfigure}[b]{0.95\linewidth}
         \centering
         \includegraphics[width=\linewidth]{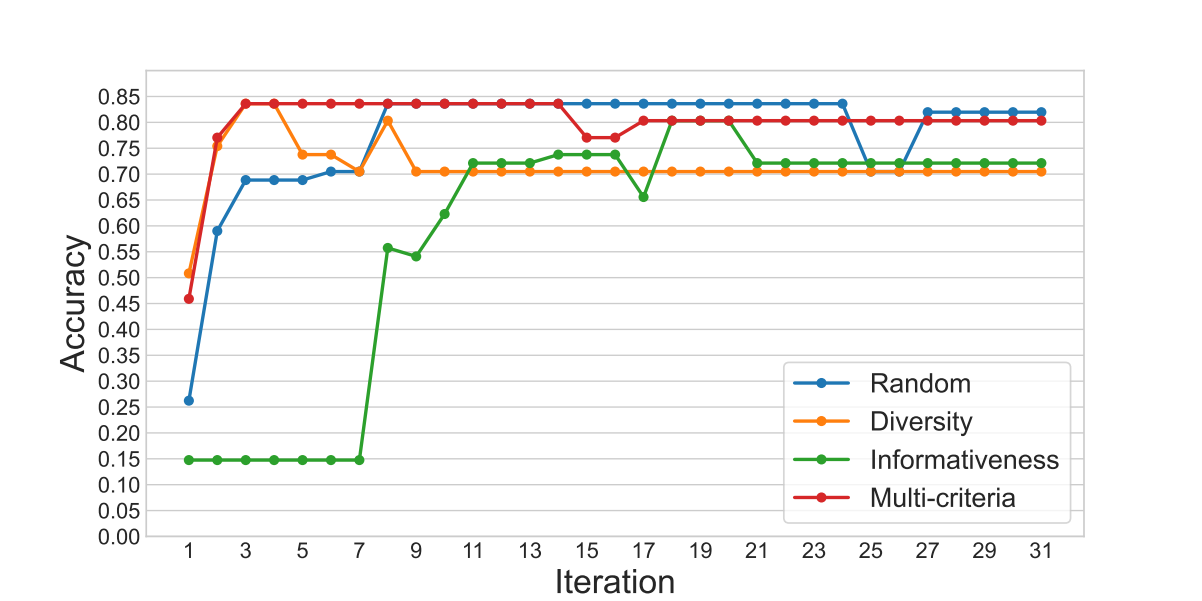}
         \caption{Comparison of accuracy.}
     \end{subfigure}
     \hfill
     \begin{subfigure}[b]{0.95\linewidth}
         \centering
         \includegraphics[width=\linewidth]{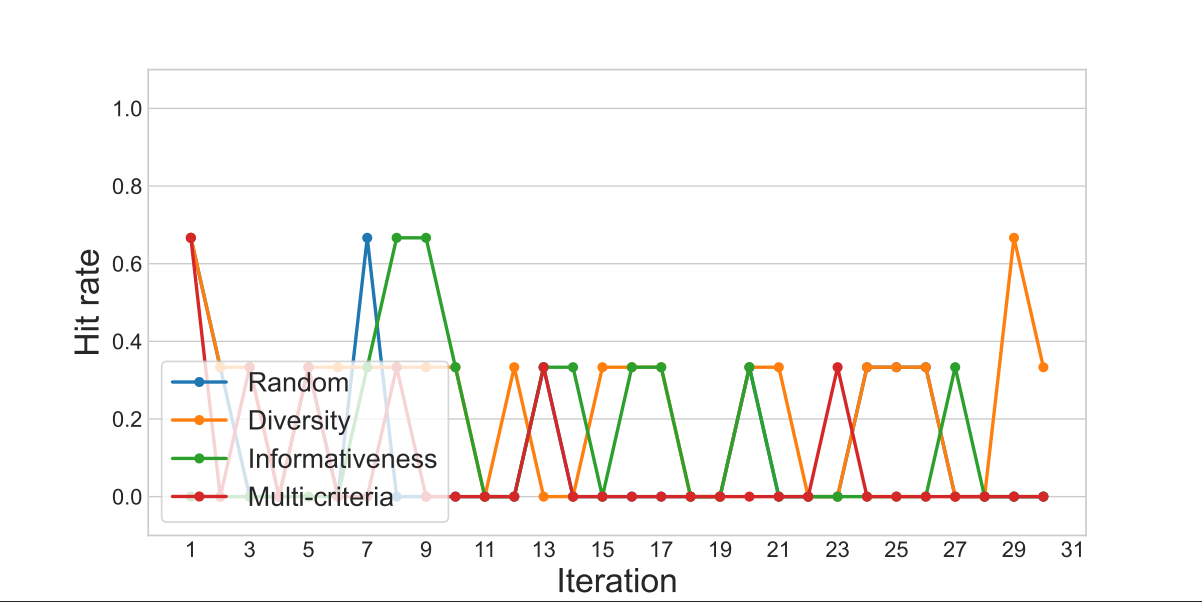}
         \caption{Comparison of hit rate.}
     \end{subfigure}
 \caption{Comparison of accuracy and hit rate in the Glaucoma diagnosis task for different data selection strategies.}
 \label{fig:selection_glaucoma}
\end{figure}

\begin{figure}[htbp]
\begin{subfigure}[b]{0.95\linewidth}
         \centering
         \includegraphics[width=\linewidth]{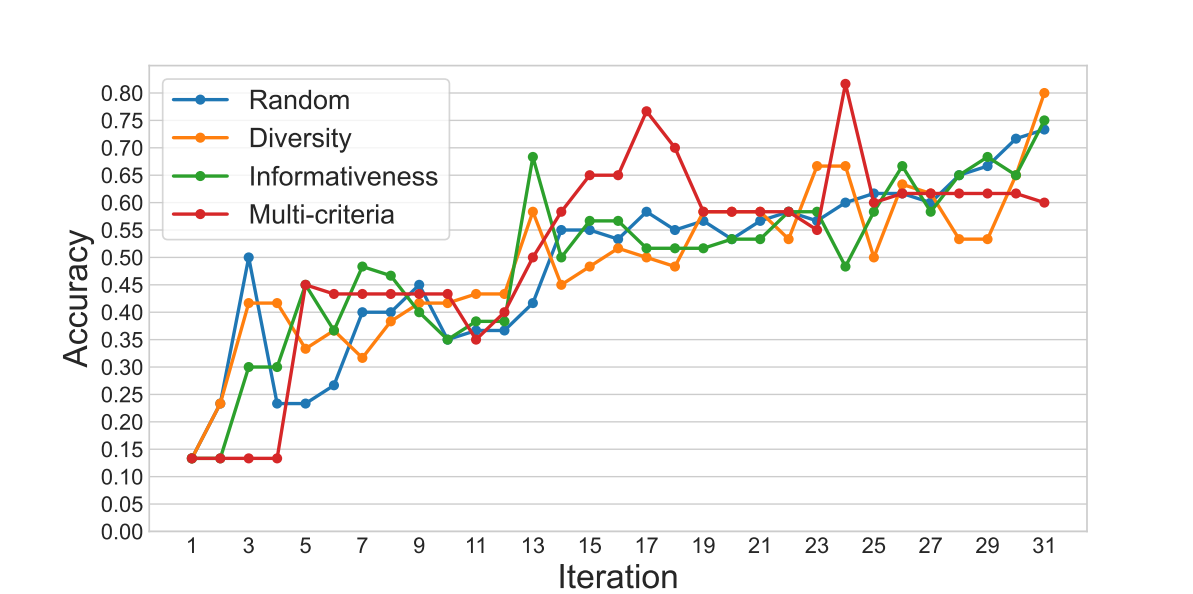}
         \caption{Comparison of accuracy.}
     \end{subfigure}
     \hfill
     \begin{subfigure}[b]{0.95\linewidth}
         \centering
         \includegraphics[width=\linewidth]{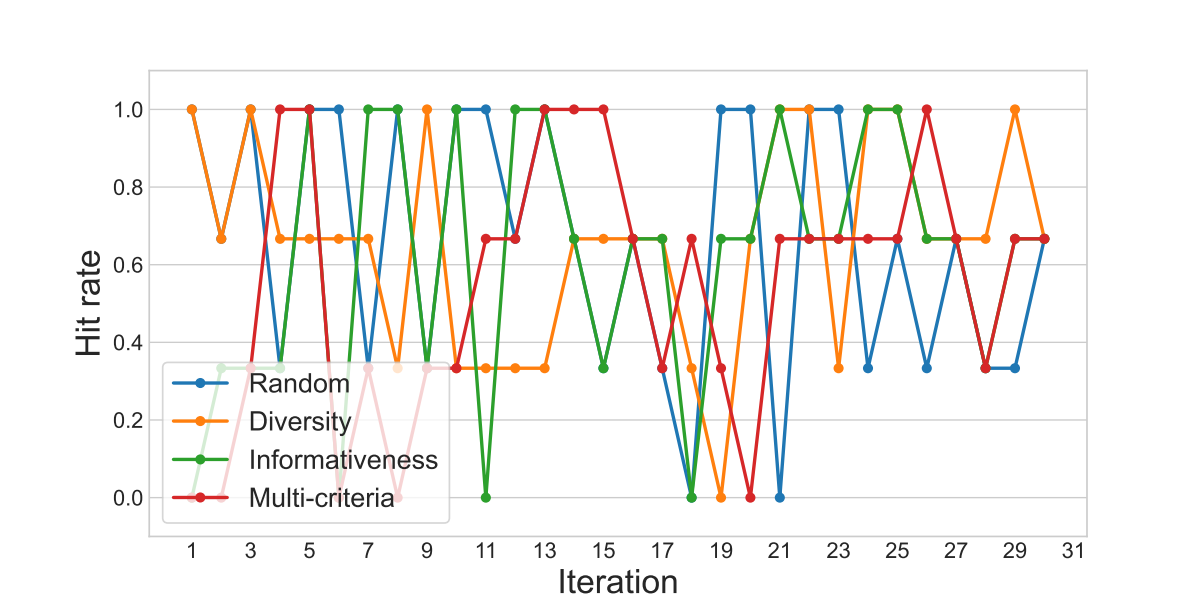}
         \caption{Comparison of hit rate.}
     \end{subfigure}
 \caption{Comparison of accuracy and hit rate in the occupation labeling task for different data selection strategies.}
 \label{fig:selection_occupation}
\end{figure}

Due to the page limit, the paper only showcases the comparison of accuracy and hit rate on the traffic scene labeling dataset for different data selection strategies. Here, we show the results in the other three tasks, bird species labeling in Figure~\ref{fig:selection_bird}, Glaucoma diagnosis in Figure~\ref{fig:selection_glaucoma}, and occupation labeling in Figure~\ref{fig:selection_occupation}.

The results match those of the traffic scene labeling task.
Our Multi-criteria achieves higher image labeling accuracy than the other three image selection strategies.
Using informativeness only, {\tool} can achieve the best hit rate when selecting images, and a high labeling accuracy as the number of iterations (i.e., the number of human-annotated images) increases.
Compared with informativeness only, our multi-criteria 

